\documentclass[a4paper,fleqn]{cas-sc}

\usepackage[utf8]{inputenc}
\usepackage[T1]{fontenc}
\usepackage{times}
\usepackage{graphicx}
\usepackage{amsmath,amssymb}
\usepackage{subcaption}
\usepackage{booktabs}
\usepackage{algorithm,algpseudocode}
\usepackage{hyperref}

\usepackage[numbers,sort&compress]{natbib}
\usepackage[section]{placeins}

\usepackage{longtable}
\usepackage{float}     
\restylefloat{figure}

\DeclareUnicodeCharacter{2248}{\ensuremath{\approx}}
\makeatletter
\def\ps@pprintTitle{%
  \let\@oddhead\@empty
  \let\@evenhead\@empty
  \let\@oddfoot\@empty
  \let\@evenfoot\@empty}
\makeatother
\hypersetup{hidelinks} 

\begin{document}

\shorttitle{}       
\shortauthors{} 
\pagestyle{plain}      
\thispagestyle{plain}  


\title [mode = title]{A Hybrid Machine Learning Approach for Synthetic Data Generation with Post Hoc Calibration for Clinical Tabular Datasets} 






\author[1]{Md Ibrahim Shikder Mahin}

\cormark[1]

\ead{mashinshikder@bubt.edu.bd}

\author[1]{ Md Shamsul Arefin}

\author[2]{Md Tanvir Hasan}

\affiliation[1]{organization={Department of Electrical \& Electronic
Engineering, Bangladesh University of Business and Technology - BUBT},
    city={Dhaka},
    country={Bangladesh}}
\affiliation[1]{organization={Department of Electrical Engineering and Computer Science, University of Michigan},
    country={United States}}

\begin{abstract}
Healthcare research and development face significant obstacles due to data scarcity and stringent privacy regulations, such as the Health Insurance Portability and Accountability Act (HIPAA) and the General Data Protection Regulation (GDPR), restricting access to essential real-world medical data. These limitations impede innovation, delay robust AI model creation, and hinder advancements in patient-centered care. Synthetic data generation offers a transformative solution by producing artificial datasets that emulate real data statistics while safeguarding patient privacy.

We introduce a novel hybrid framework for high-fidelity healthcare data synthesis integrating five augmentation methods: noise injection, interpolation, Gaussian Mixture Model (GMM) sampling, Conditional Variational Autoencoder (CVAE) sampling, and SMOTE, combined via a reinforcement learning-based dynamic weight selection mechanism. Its key innovations include advanced calibration techniques moment matching, full histogram matching, soft and adaptive soft histogram matching, and iterative refinement that align marginal distributions and preserve joint feature dependencies.

Evaluated on the Breast Cancer Wisconsin (UCI Repository) and Khulna Medical College cardiology datasets, our calibrated hybrid achieves Wasserstein distances as low as 0.001 and Kolmogorov–Smirnov statistics around 0.01, demonstrating near-zero marginal discrepancy. Pairwise trend scores surpass 90\%, and Nearest Neighbor Adversarial Accuracy approaches 50\%, confirming robust privacy protection. Downstream classifiers trained on synthetic data achieve up to 94\% accuracy and F1 scores above 93\%, comparable to models trained on real data. This scalable, privacy-preserving approach matches state-of-the-art methods, sets new benchmarks for joint-distribution fidelity in healthcare, and supports sensitive AI applications.
\end{abstract}


\maketitle
\section{Introduction}
Data-driven healthcare depends on access to high-quality, representative datasets for building and validating robust AI models \cite{1}. In practice, U.S. and European privacy regulations impose strict limits on the use, sharing, and secondary processing of personal health information, which constrains access to real-world data for model development \cite{2,3}. As a mitigation, Bourou et al. \cite{4} highlight synthetic data generation as a practical pathway that can mirror key statistics of real datasets while reducing disclosure risk. In this study, we focus on oncology and cardiology contexts using the Breast Cancer Wisconsin (Original) and Breast Cancer Wisconsin (Diagnostic) datasets from the UCI Machine Learning Repository, together with a heart-disease cohort from the Department of Cardiology, Khulna Medical College \cite{5}.

Chen et al. and related efforts show that synthetic data can enable analyses and workflow simulations under data-access constraints while protecting identities \cite{6} \cite{7}. In discrete EHR settings, Choi et al. introduced \textit{medGAN} to synthesize multi-label records for predictive modeling in scarce-data regimes \cite{8}. In medical imaging, Frid-Adar et al. demonstrated that GAN-based augmentation improves diagnostic performance with limited samples\cite{9}. Beyond these exemplars, Yelmen et al. illustrate privacy-preserving potential in genomics, underscoring the broader promise of synthetic data for ethical, regulation-aware AI in biomedicine\cite{10}.

Still, as emphasized by prior work \cite{11}, generating high-fidelity \emph{tabular} healthcare data that preserves complex dependencies is challenging. Classical probabilistic models often struggle with high dimensionality and nonlinearity \cite{12}. While GAN-based approaches advance realism \cite{13} and variational autoencoders provide a complementary probabilistic route \cite{14}, practical deployment can be hindered by mode collapse, overfitting, and limited training data \cite{15}. Consequently, many pipelines rely on post-generation calibration to retain downstream utility \cite{16}.

Hybrid strategies have emerged to combine complementary strengths across generators and augmentations \cite{17}. In healthcare records, Torfi and Fox integrated correlation-capturing techniques with GANs to improve realism \cite{18}. For post-hoc correction, Deville and Särndal introduced calibration estimators to align sample moments with real-data targets \cite{40}, and Bourou et al. discussed histogram-based calibration as a practical tool to reduce distributional drift \cite{4}; related work formalizes moment and histogram matching for robust alignment \cite{19,20}. Despite this progress, gaps often persist between generation and calibration stages, motivating unified workflows tailored to healthcare tabular data \cite{21}.

Evidence across domains reinforces a hybrid-and-calibrate paradigm. In imaging, Frid-Adar et al. enriched rare phenotypes via adversarial augmentation to improve lesion classification  \cite{26}. In tabular settings, studies have applied synthetic data to bolster fraud detection through minority-class expansion \cite{27} and have leveraged generative models for market simulations in finance \cite{28}. Within healthcare records, Rahman et al. employed VAE-based frameworks to produce clinically meaningful cohorts, while latent-variable formulations have also been explored for financial time series  \cite{29}\cite{30}. Conditional models further increase controllability by incorporating labels and clinical context \cite{31}. At the same time, practical adoption must account for computational demands and the risk of biased generations under limited or skewed training data \cite{32}.

Among augmentation tools, Mousavi et al. showed that noise injection can mitigate class imbalance and improve generalization in low-resource regimes \cite{33}. Interpolation-based oversampling such as SMOTE remains a simple, effective mechanism for expanding minority classes \cite{34}; Li et al. proposed adaptive interpolation for rare-disease classification \cite{35}, and subsequent work reported gains in fraud analytics \cite{36}, though linearity assumptions can break down in high-dimensional or nonlinear spaces \cite{37}. To capture multi-modality, Wang et al. used Gaussian Mixture Models (GMMs) for cluster-aware sampling \cite{38}, while noting challenges from sparsity and the curse of dimensionality in complex clinical tables \cite{39}. On the calibration front, adaptive and iterative histogram strategies further refine alignment through dynamic, successive adjustments \cite{41}. Compared with imaging, calibration methods purpose-built for healthcare tabular data remain comparatively underexplored \cite{39}. Open-source tooling has lowered barriers to adoption \cite{4}, cohort-level engines have demonstrated feasibility for patient-centered outcomes research \cite{6}, synthetic imaging has expanded rare categories for model development \cite{42}, and probabilistic–deep hybrids for clinical time series (e.g., Du et al.) provide additional evidence for combining model families \cite{43}. The benchmark datasets employed here serve as established testbeds for assessing fidelity, utility, and privacy safeguards in oncology and cardiology \cite{5}.

Even so, many current methods struggle to capture higher-order dependencies essential for clinical inference in high-dimensional tables \cite{44}, and the absence of standardized evaluation protocols complicates fair comparisons across methods and datasets \cite{45}. Motivated by these gaps, we ask whether a unified hybrid framework that integrates multiple augmentation techniques with targeted calibration can produce high-fidelity synthetic data while preserving downstream utility on real healthcare tasks \cite{22}. We hypothesize that combining noise injection, interpolation, GMM sampling, Conditional Variational Autoencoder (CVAE) sampling, and SMOTE followed by moment- and histogram-based calibration will more closely match real distributions than single-method baselines \cite{23}. We evaluate distributional similarity using Wasserstein distance and the Kolmogorov–Smirnov statistic \cite{24} and benchmark against widely used tabular synthesizers in the Synthetic Data Vault ecosystem \cite{25}. The remainder of this article is organized as follows: Section~1 reviews related work, Section~2 details the methodology, Section~3 presents experiments and analysis, and Section~4 discusses implications, limitations, and future directions.
\section{Methodology}
This research introduces a comprehensive framework for generating high-fidelity synthetic data through a hybrid model that integrates multiple data augmentation techniques within a unified pipeline. The framework addresses data scarcity and privacy concerns in sensitive domains by combining the strengths of various approaches to capture diverse aspects of data distributions. Additionally, it tackles the significant challenge of limited real-world data availability for training robust models. Real-world data collection is often costly, time-consuming, and constrained by privacy regulations, especially in healthcare, finance, and other sensitive sectors. Furthermore, real-world datasets frequently suffer from imbalances and insufficient representation of rare events, which can severely limit model performance. The hybrid model comprises two main components: (i) synthesis of new data samples using a combination of noise injection, interpolation, Gaussian Mixture Modeling (GMM), Conditional Variational Autoencoder (CVAE), and SMOTE-like interpolation; and (ii) a series of calibration procedures that refine the synthetic data distribution to closely align with the original data.

\begin{figure}[ht]
    \centering
    \includegraphics[width=\textwidth]{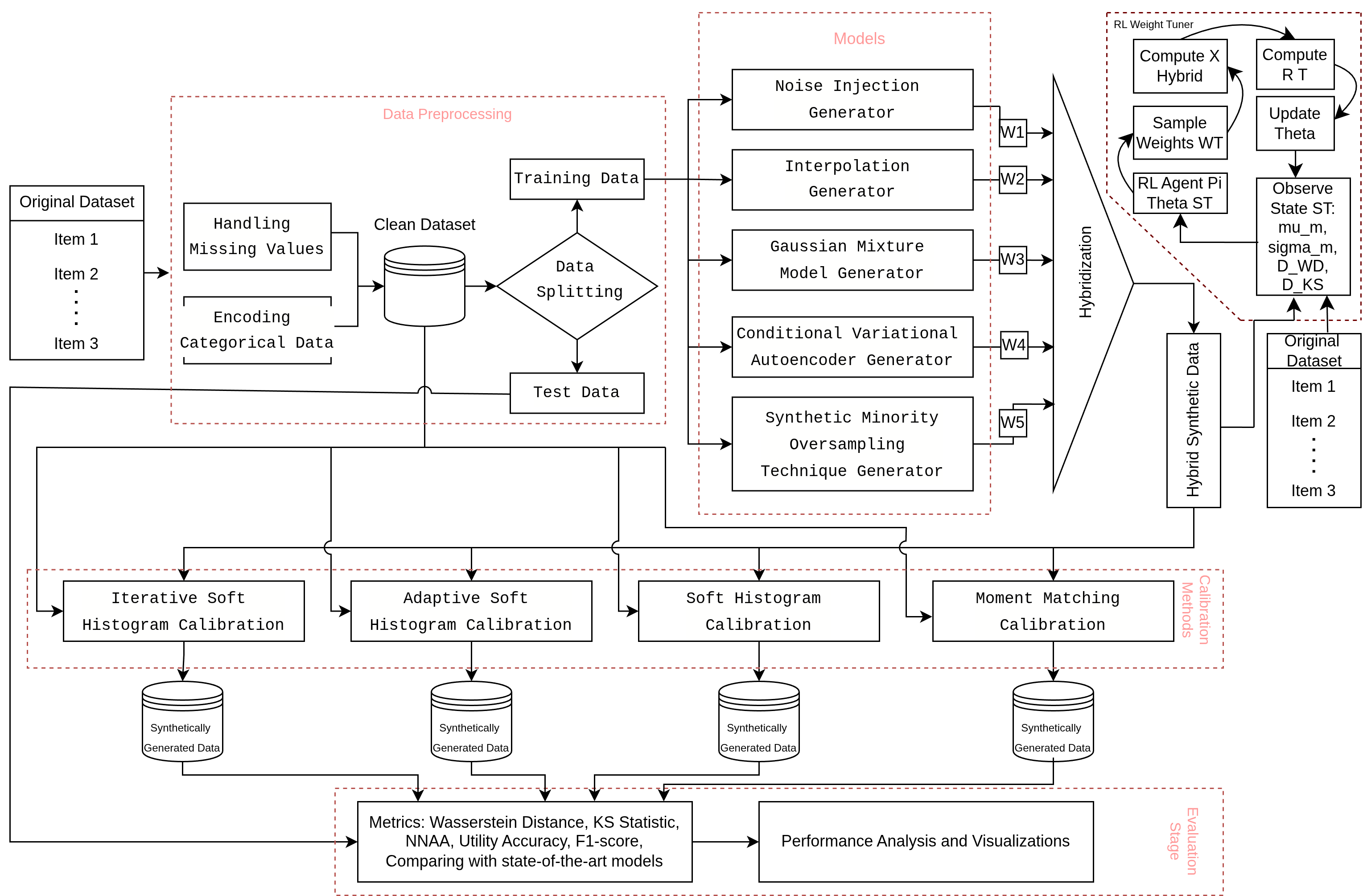}
    \caption{\textit{Workflow of the hybrid synthetic data generation framework. The figure illustrates data preprocessing, hybrid generation via five techniques, sequential application of the five calibration methods, and evaluation stages. The five post-calibration methods are depicted as iterative refinements to align synthetic distributions with original data.}}
    \label{fig:framework}
\end{figure}

Figure \ref{fig:framework} provides a visual representation of the overall hybrid model workflow, illustrating how the various components interact throughout the synthetic data generation process. The figure depicts the sequential stages of the framework, beginning with data preprocessing, followed by individual synthetic data generation through multiple techniques, hybridization through weighted averaging, and concluding with calibration methods to enhance data fidelity.

\subsection{Datasets}
Three distinct datasets were utilized to validate the efficacy of the proposed hybrid synthetic data generation model. Two datasets pertain to breast cancer, sourced from the University of California, Irvine (UCI) Machine Learning Repository \cite{dua2019uci}, while the third dataset relates to cardiovascular disease (CVD), collected from patients at the Department of Cardiology, Khulna Medical College, Bangladesh.

The first breast cancer dataset, referred to as the \textbf{Diagnostic Breast Cancer Dataset}, consists of 569 samples with 32 attributes including patient ID, diagnosis (malignant or benign), and various quantitative features extracted from breast mass imagery. The second breast cancer dataset, termed \textbf{Original Breast Cancer Dataset}, comprises 699 instances with 11 attributes, detailing cell characteristics associated with malignancy.

The cardiovascular disease dataset contains 300 records with 14 attributes, capturing demographic information, clinical metrics such as blood pressure and cholesterol levels, and binary indicators for the presence or absence of cardiovascular conditions.

\begin{table}[ht]
\centering
\caption{Summary of datasets used in the study}
\label{tab:datasets}
\begin{tabular}{lcc}
\toprule
Dataset & Instances & Attributes \\
\midrule
Diagnostic Breast Cancer & 569 & 31 \\
Original Breast Cancer & 699 & 10 \\
Cardiovascular Disease & 300 & 13 \\
\bottomrule
\end{tabular}
\end{table}

\subsection{Data Preprocessing}
The preprocessing stage ensures data quality and compatibility with subsequent model components. This involves handling missing values, categorical features, and data splitting.

The input dataset is split into training and testing sets using a stratified approach to maintain class distributions across subsets: 
\begin{equation}
\mathcal{D} = \{\mathbf{x}_1, \mathbf{x}_2, \ldots, \mathbf{x}_N\} \rightarrow \mathcal{D}_{\text{train}}, \mathcal{D}_{\text{test}}
\end{equation}
where $\mathcal{D}$ represents the entire dataset, and $\mathcal{D}_{\text{train}}$ and $\mathcal{D}_{\text{test}}$ are the resulting training and testing sets, respectively \cite{hastie2009elements}. The stratified splitting ensures that each subset maintains representative class distributions, which is crucial for maintaining model performance across different classes. Proper data splitting prevents overfitting and allows for unbiased evaluation of model performance on unseen data.

Missing values are addressed through imputation, typically replacing NaN values with a neutral value (e.g., zero) or a statistically informed estimate: 
\begin{equation}
\mathbf{x}_i^{\text{imputed}} = 
\begin{cases} 
\mathbf{x}_i & \text{if } \mathbf{x}_i \text{ is not missing} \\
\mu_{\text{feature}} & \text{if } \mathbf{x}_i \text{ is missing}
\end{cases}
\end{equation}
where $\mu_{\text{feature}}$ represents the mean or median of the corresponding feature \cite{rubin1987statistical}. Proper handling of missing values is essential to prevent bias and maintain data integrity, which directly impacts the quality of the synthetic data generated. Missing data can introduce systematic errors if not properly addressed, potentially leading to models that perform poorly on real-world data.

Categorical labels are encoded into a numerical format suitable for machine learning algorithms: 
\begin{equation}
\mathbf{y}_i^{\text{encoded}} = \text{OneHot}(\mathbf{y}_i)
\end{equation}
where $\text{OneHot}(\cdot)$ denotes the one-hot encoding function \cite{bishop2006pattern}. Encoding categorical variables allows machine learning models to properly interpret and utilize these features during training. Without proper encoding, models may misinterpret categorical data as ordinal, leading to incorrect learning and reduced performance.

\subsection{Hybrid Synthetic Data Generation}
The core of the proposed methodology lies in the ensemble of diverse synthetic data generators, each utilizing distinct statistical and machine learning techniques.

The Noise Injection technique introduces Gaussian noise with a controlled level to original data samples: 
\begin{equation}
\mathbf{x}_i^{\text{noisy}} = \mathbf{x}_i + \epsilon, \quad \epsilon \sim \mathcal{N}(0, \sigma^2 \mathbf{I})
\end{equation}
where $\sigma$ controls the noise intensity \cite{goodfellow2016deep}. Noise injection helps to increase data diversity and can improve model robustness by creating perturbed versions of existing data points. This technique is particularly valuable in scenarios where data is limited, as it creates variations that help models generalize better to unseen data. The controlled noise level ensures that the perturbations are sufficient to enhance diversity without introducing unrealistic data points.

Interpolation generates new samples by linearly interpolating between an original data point and a randomly selected data point of the same class:
\begin{equation}
\mathbf{x}_i^{\text{interp}} = \lambda \mathbf{x}_i + (1 - \lambda) \mathbf{x}_j, \quad \lambda \sim \text{Uniform}(0, 1)
\end{equation}
where $\mathbf{x}_j$ is a randomly selected data point from the same class as $\mathbf{x}_i$ \cite{devries2017improved}. Interpolation helps to create new data points within the existing data manifold, capturing local patterns and relationships between data points. This technique is particularly effective for generating synthetic data that maintains the underlying structure of the original data, making it suitable for applications where data geometry is important.

Class-conditional Gaussian Mixture Model (GMMs) are trained on the training data, modeling each class as a mixture of Gaussian components: 
\begin{equation}
\mathbf{x}_i^{\text{GMM}} \sim \sum_{k=1}^K \pi_k \mathcal{N}(\boldsymbol{\mu}_k, \boldsymbol{\Sigma}_k)
\end{equation}
where $\pi_k$, $\boldsymbol{\mu}_k$, and $\boldsymbol{\Sigma}_k$ are the mixture weights, means, and covariances for the $k$-th Gaussian component \cite{bishop2006pattern}. GMMs are powerful for capturing complex multimodal distributions, allowing the generation of synthetic data that reflects the underlying statistical structure of each class. This technique is particularly valuable for datasets with complex distributions that cannot be adequately modeled by simpler techniques.

A Conditional Variational Autoencoder (CVAE) is trained to learn a latent space representation conditioned on class labels: 
\begin{equation}
\mathbf{z}_i \sim \mathcal{N}(\boldsymbol{\mu}_i, \boldsymbol{\Sigma}_i), \quad \mathbf{x}_i^{\text{CVAE}} = \text{Decoder}(\mathbf{z}_i, y_i)
\end{equation}
where $\mathbf{z}_i$ is a latent vector sampled from the encoder's output distribution, and $y_i$ is the class label \cite{kingma2014auto}. CVAEs are particularly effective for generating high-quality synthetic samples by learning complex, non-linear data distributions in a lower-dimensional latent space. This technique is especially valuable for high-dimensional data where traditional methods may struggle to capture the underlying structure.

Synthetic Minority Over-sampling Technique (STOME) creates synthetic examples for each original sample by linear interpolation with its nearest neighbor in the same class:
\begin{equation}
\mathbf{x}_i^{\text{STOME}} = \mathbf{x}_i + \gamma (\mathbf{x}_j - \mathbf{x}_i), \quad \gamma \sim \text{Uniform}(0, 1)
\end{equation}
where $\mathbf{x}_j$ is the nearest neighbor of $\mathbf{x}_i$ within the same class \cite{chawla2002smote}. STOME is particularly useful for addressing class imbalance by generating synthetic examples for minority classes, ensuring that the synthetic dataset maintains proper class representation. This technique helps prevent models from being biased toward majority classes, which is crucial for fair and accurate performance across all classes.

\subsection{Hybridization Stage: RL‐Driven Weight Selection}
\label{sec:rl_weight_selection}

As illustrated in Figure \ref{fig:framework}, our hybrid model integrates an RL‐based weight tuning module within the overall architecture.  In particular, the dashed inset in the top right of Figure \ref{fig:framework} depicts the RL agent observing distributional statistics and producing a weight vector, which then modulates the convex combination of generator outputs.

We cast the hybridization step as a Markov decision process (MDP) in which an RL agent learns to assign weights to the $M$ individual generators, thereby adapting to complex distributional discrepancies.  At each time step $t$, the agent observes a state
\begin{equation}
s_t = \Bigl\{\mu_m,\;\sigma_m\Bigr\}_{m=1}^M\;\cup\;\bigl\{D_{\mathrm{WD}},\;D_{\mathrm{KS}}\bigr\},
\end{equation}
where $\mu_m$ and $\sigma_m$ are the empirical mean and standard deviation of samples from generator $m$, and $D_{\mathrm{WD}}$ and $D_{\mathrm{KS}}$ denote the global Wasserstein and Kolmogorov–Smirnov divergences between the current hybrid output and the real data distribution.

The agent’s action is a weight vector
\[
a_t = w_t = [\,w_{t,1},\,\dots,\,w_{t,M}\,],
\qquad
\sum_{m=1}^M w_{t,m} = 1,
\]
sampled from a stochastic policy $\pi_\theta(a_t \mid s_t)$ parameterized by $\theta$.  The resulting hybrid sample is computed as
\begin{equation}
x_i^{\mathrm{hyb}} \;=\; \sum_{m=1}^M w_{t,m}\,x_i^{(m)},
\qquad
\sum_{m=1}^M w_{t,m}=1.
\tag{9$^\prime$}
\end{equation}

To guide learning, we define the per‐step reward as the negative average calibration loss across $d$ features,
\begin{equation}
r_t \;=\;-\,\frac{1}{d}\sum_{j=1}^d 
\mathrm{WD}\bigl(x^{\mathrm{hyb}}_{:,j},\,x^{\mathrm{real}}_{:,j}\bigr)\,,
\end{equation}
so that maximizing cumulative reward directly minimizes distributional discrepancies.

Policy parameters $\theta$ are updated via the REINFORCE rule:
\begin{equation}
\theta\;\leftarrow\;\theta\;+\;\alpha\;\nabla_\theta\log\pi_\theta\bigl(w_t\mid s_t\bigr)\bigl(r_t - b\bigr),
\tag{10}
\end{equation}
where $\alpha$ is the learning rate and $b$ is a learned baseline used to reduce gradient variance.  In practice, we implement $\pi_\theta$ as a two‐layer neural network with softmax output to ensure $\sum_m w_{t,m}=1$, and learn $b$ via an exponential moving average of past rewards.

This RL‐based approach outperforms static equal‐weight hybrids by dynamically emphasizing generators that reduce calibration error in real time, yielding higher pairwise trend fidelity and downstream utility gains in classification tasks.  Moreover, by framing weight selection as an MDP, our method naturally accommodates extensions such as off‐policy correction or actor–critic variants for even greater stability and sample efficiency.

\subsection{Calibration Methods Stage}
The calibration stage adjusts the statistical properties of the synthetic data to align more closely with the original training data.

Moment Matching Calibration technique adjusts the mean and standard deviation of each feature in the hybrid synthetic data: 
\begin{equation}
\mathbf{x}_i^{\text{calibrated}} = \alpha (\mathbf{x}_i^{\text{hybrid}} - \boldsymbol{\mu}_{\text{synth}}) + \boldsymbol{\mu}_{\text{orig}}, \quad \alpha = \frac{\boldsymbol{\sigma}_{\text{orig}}}{\boldsymbol{\sigma}_{\text{synth}}}
\end{equation}
where $\boldsymbol{\mu}_{\text{synth}}$ and $\boldsymbol{\sigma}_{\text{synth}}$ are the mean and standard deviation of the synthetic data, and $\boldsymbol{\mu}_{\text{orig}}$ and $\boldsymbol{\sigma}_{\text{orig}}$ are those of the original data \cite{pearson1895contributions}. Moment matching is computationally efficient and helps to align the first two statistical moments of the distributions, which are crucial for data representation and model performance. Proper alignment of these moments ensures that the synthetic data maintains similar scale and central tendency as the original data, preventing models from learning incorrect patterns.

Full Histogram Matching Calibration method matches the entire histogram distribution of each feature: 
\begin{equation}
\mathbf{x}_i^{\text{calibrated}} = \text{HistogramMatch}(\mathbf{x}_i^{\text{hybrid}}, \text{orig\_histogram})
\end{equation}
where $\text{HistogramMatch}(\cdot)$ denotes the histogram matching function \cite{gonzalez2008digital}. Full histogram matching aims to achieve a more comprehensive distributional alignment than moment matching, capturing higher-order statistical properties that can significantly impact model training. By matching the entire histogram, this technique ensures that the synthetic data not only has similar mean and variance but also similar shape and modality as the original data.

Soft Histogram Matching Calibration approach blends the hybrid synthetic data with the full histogram-matched synthetic data: 
\begin{equation}
\mathbf{x}_i^{\text{calibrated}} = \alpha \mathbf{x}_i^{\text{hybrid}} + (1 - \alpha) \mathbf{x}_i^{\text{full\_match}}
\end{equation}
where $\alpha$ is a fixed blending factor (e.g., 0.5) \cite{reinhard2001color}. Soft histogram matching provides a compromise between preserving the diversity from the hybridization stage and benefiting from histogram alignment, maintaining a balance between data fidelity and representativeness. This technique helps to avoid over-calibration while still improving the statistical alignment with the original data.

Adaptive Soft Histogram Matching Calibration method extends soft histogram matching by using adaptive alpha values for each feature: 
\begin{equation}
\alpha_d = \frac{1}{1 + \exp(-\beta (D_d - \tau))}, \quad \mathbf{x}_{i,d}^{\text{calibrated}} = \alpha_d \mathbf{x}_{i,d}^{\text{hybrid}} + (1 - \alpha_d) \mathbf{x}_{i,d}^{\text{full\_match}}
\end{equation}
where $D_d$ represents the distributional discrepancy for feature $d$, $\beta$ controls the slope of the sigmoid function, and $\tau$ is a threshold parameter \cite{kingma2015adam}. Adaptive soft histogram matching allows for more nuanced calibration by applying stronger histogram matching to features with greater distributional discrepancies, ensuring that each feature is appropriately aligned with the original data. This feature-specific calibration helps to address the varying levels of discrepancy across different features, improving overall data quality.

Iterative Soft Histogram Matching Calibration technique iteratively refines the soft histogram matching process: 
\begin{equation}
\mathbf{x}_i^{(t+1)} = \alpha^{(t)} \mathbf{x}_i^{(t)} + (1 - \alpha^{(t)}) \mathbf{x}_i^{\text{full\_match}}, \quad \alpha^{(t)} = \frac{W^{(t)}}{W^{(t)} + \epsilon}
\end{equation}
where $W^{(t)}$ is the Wasserstein distance at iteration $t$, and $\epsilon$ is a small constant to prevent division by zero \cite{cuturi2013sinkhorn}. The iterative refinement process allows for a more refined and potentially optimal calibration, aiming to minimize the distributional divergence between the synthetic and original data. This technique ensures that the calibration process converges to a solution where the synthetic data is as close as possible to the original data in terms of statistical properties.

\subsection{Evaluation of Synthetic Data}
The quality, utility, and privacy of the synthetic data are evaluated through several complementary metrics integrated within the framework:
The synthetic data is assessed for how well its distribution $Q$ approximates the original data distribution $P$:
\begin{itemize}
    \item \textbf{Wasserstein Distance:}  
    \begin{equation}
        W(P,Q) = \inf_{\gamma \in \Gamma(P,Q)} \mathbb{E}_{(x,y) \sim \gamma} [\| x - y \|]
    \end{equation}
    where $\Gamma(P,Q)$ is the set of all joint distributions with marginals $P$ and $Q$ \cite{rubner2000earth}.
    
    \item \textbf{Kolmogorov-Smirnov Statistic:}  
    \begin{equation}
        KS(P,Q) = \sup_{x} \left| F_P(x) - F_Q(x) \right|
    \end{equation}
    with $F_P(x)$ and $F_Q(x)$ being the cumulative distribution functions (CDFs) of the original and synthetic data, respectively \cite{smirnov1948table}.
\end{itemize}
These metrics provide insight into whether the synthetic data maintains the key statistical properties of the original dataset. 

Nearest Neighbor Adversarial Accuracy (NNAA) employs a 1-nearest neighbor classifier to distinguish between synthetic and real samples. Nearest Neighbor Adversarial Accuracy (NNAA) is calculated as:
\begin{equation}
    \mathrm{NNAA} = \frac{1}{N} \sum_{i=1}^{N} \mathbb{I}\bigl[\hat{y}_i = y_i\bigr]
\end{equation}
where $\hat{y}_i$ is the predicted label, $y_i$ is the true label, and $N$ is the number of samples \cite{pedregosa2011scikit}. An NNAA value approaching 50\% suggests that the synthetic data is nearly indistinguishable from the real data.

Utility Evaluation quantifies the usefulness of the synthetic data by training machine learning models on it and evaluating their performance on the original test set. Two key performance metrics under Utility Evaluation are:
\begin{itemize}
    \item \textbf{Classification Accuracy:}
    \begin{equation}
        \mathrm{Acc} = \frac{\text{Number of Correct Predictions}}{\text{Total Number of Predictions}}
    \end{equation}
    \item \textbf{Weighted F1 Score:}
    \begin{equation}
        \mathrm{F1} = \frac{2 \times \text{Precision} \times \text{Recall}}{\text{Precision} + \text{Recall}}
    \end{equation}
\end{itemize}
These metrics validate that models trained on synthetic data generalize effectively to real-world data \cite{chen2016xgboost}.

\subsubsection{Benchmarking Against Established Methods}  
To contextualize the performance of the proposed framework, the synthetic data quality is benchmarked against well-established generators available through the Synthetic Data Vault (SDV) framework. Specifically, the following methods are used as baselines: CTGAN \cite{xu2019modeling}, TVAE \cite{xu2019modeling2}, Gaussian Copula \cite{reiter1990using}, and CopulaGAN \cite{yoon2019time}. The SDV evaluation protocol provides detailed quality reports that assess metrics such as column shape preservation and feature pair trends, resulting in an overall synthesis quality score. These benchmarks help demonstrate that our hybrid approach effectively balances data fidelity, utility, and privacy.

Overall, the integrated evaluation strategy within our framework provides continuous feedback to ensure that the synthetic data not only replicates the underlying statistical characteristics of the original data but is also practical for downstream predictive tasks and resilient against privacy attacks.

\section{Experimental Results}
We applied six hybrid generation methods \emph{Raw Hybrid}, \emph{Moment Matching}, \emph{Full Histogram}, \emph{Soft Histogram (alpha=0.5)}, \emph{Adaptive Soft Histogram}, and \emph{Iterative Soft Histogram} to each data set. The methods vary in how they capture distributional properties: for instance, \textit{Moment Matching} preserves feature means and variances, \textit{Full Histogram} preserves each feature's full empirical distribution, and \textit{Iterative Soft Histogram} incrementally adjusts distributions to match the target. We assess how these methods balance distribution fidelity against downstream utility. 

In the following, we first present results on the Breast Cancer Wisconsin (Original) dataset, then on the Breast Cancer Wisconsin (Diagnostic) dataset, and finally on the Cardiovascular Disease dataset. Figures and tables are referenced for each case to illustrate the comparative performance.

\subsection{Breast Cancer Wisconsin (Original) Dataset Analysis}
The Breast Cancer Wisconsin (Original) dataset is a classical binary classification dataset with 10 continuous features (e.g., clump thickness, uniformity of cell size) plus a class label. We analyze the quality of the synthetic data generated for this dataset by examining dimensionality reduction plots (PCA, t-SNE, UMAP), correlation heatmaps, and feature density distributions. We also assess distribution fidelity metrics and the utility of the synthetic data for downstream classification, paying special attention to how calibration methods improved the results.

\subsubsection{Dimensionality Reduction Visualizations (PCA, t-SNE, UMAP)}
\begin{figure}[ht]
  \centering
  \begin{subfigure}{0.49\textwidth}
    \includegraphics[width=\linewidth]{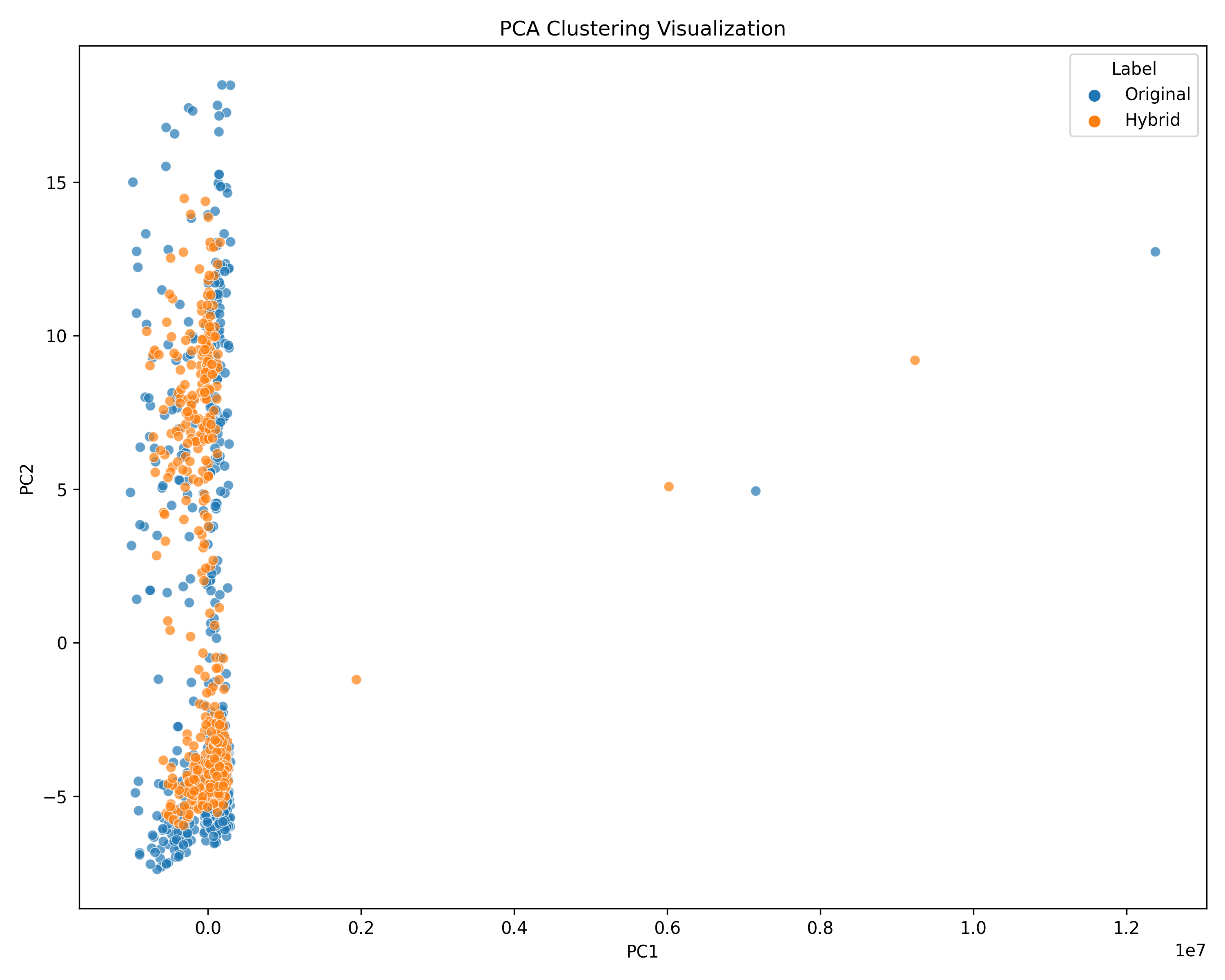}
    \caption{PCA (2 components)}
    \label{fig:bc-original-pca}
  \end{subfigure}
  \begin{subfigure}{0.49\textwidth}
    \includegraphics[width=\linewidth]{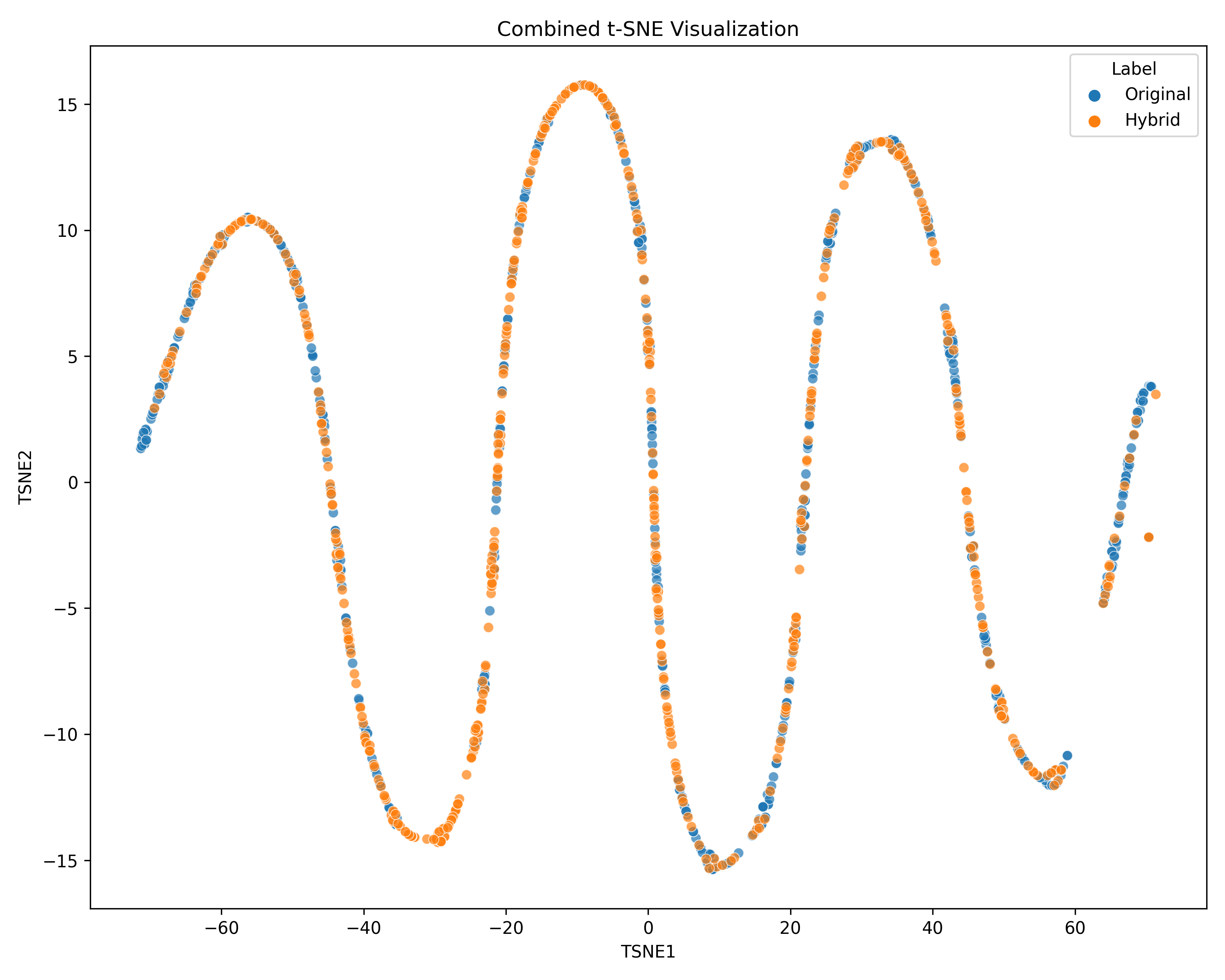}
    \caption{t-SNE}
    \label{fig:bc-original-tsne}
  \end{subfigure}
  \begin{subfigure}{0.49\textwidth}
    \includegraphics[width=\linewidth]{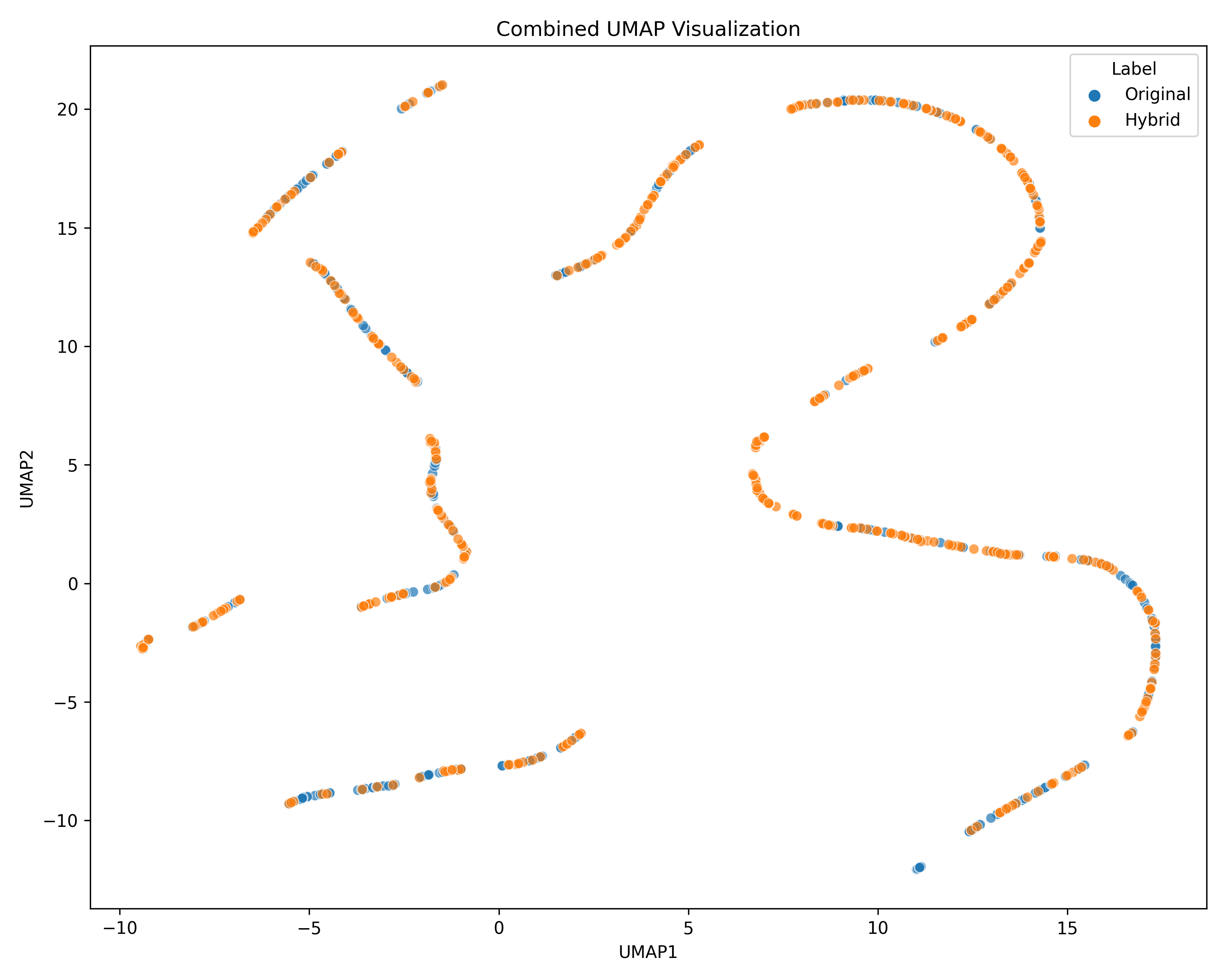}
    \caption{UMAP}
    \label{fig:bc-original-umap}
  \end{subfigure}
  \caption{Projection of real vs.\ synthetic Breast Cancer (Original) data in PCA, t-SNE, and UMAP spaces. Real data points and synthetic data points are plotted together to visualize overlap; ideally, synthetic points align with the real data distribution in each projection.}
  \label{fig:bc-original-proj}
\end{figure}

Real data points and synthetic data points are plotted together to visualize their overlap in each projection (Figure~\ref{fig:bc-original-proj}). Ideally, the calibrated synthetic points should align with the real data distribution in each low-dimensional view. In these plots, the synthetic data indeed closely overlaps the real data. The PCA plot (Figure~\ref{fig:bc-original-proj}a) shows that the first two principal components capture a large portion of variance, and the synthetic distribution covers the same span in PC space as the real distribution. Any subtle principal-component-level differences are minimal; for instance, the range of PC1 values and the clustering along PC2 for synthetic data nearly mirror the real data's PCA projection. The t-SNE visualization (Figure~\ref{fig:bc-original-proj}b) indicates that the generative model has learned the underlying class structure synthetic malignant cases occupy the same region as real malignant cases, and similarly for benign cases. There are no significant synthetic ``ghost'' clusters (spurious groupings with no real counterpart); the synthetic data covers the multimodal structure without introducing extra modes. The UMAP plot (Figure~\ref{fig:bc-original-proj}c) likewise demonstrates that synthetic samples fill in the manifold of real data. We observe that any subtle gaps between real and synthetic points in the uncalibrated output have been minimized after applying calibration. For example, some extreme or outlier points present in the real data were initially under-represented in the raw synthetic data, but after calibration these points appear in the synthetic set, reducing divergence between the two distributions. Overall, the dimensionality reduction visualizations confirm that the calibrated synthetic data closely reproduces the global and local structure of the Breast Cancer (Original) dataset.

\subsubsection{Marginal and Joint Distribution Comparison}
\begin{figure}[ht]
\centering
\includegraphics[width=1\textwidth,height=0.8\textheight,keepaspectratio]{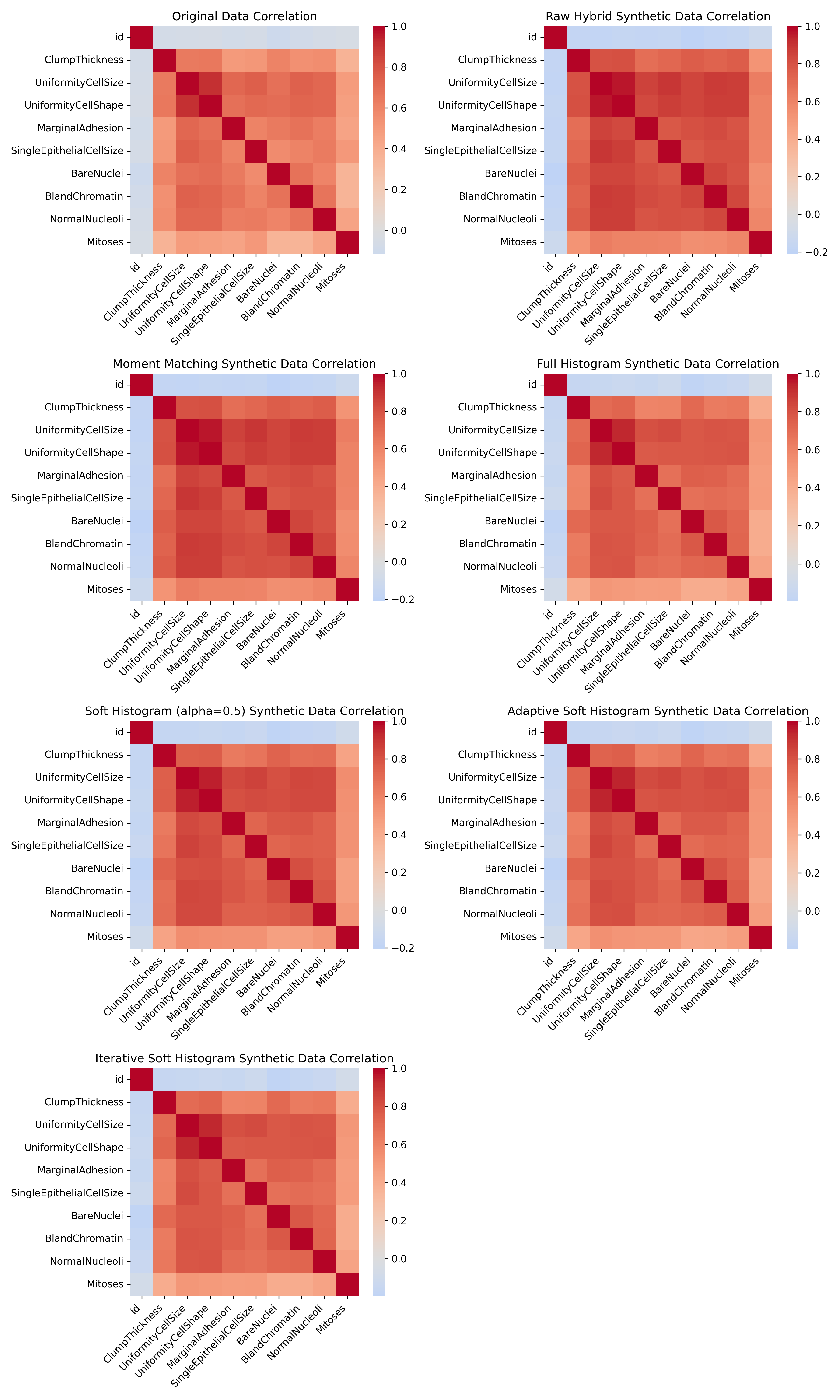}
\caption{Correlation matrices of features for the Breast Cancer (Original) dataset: (left) real data, (right) synthetic data (post-calibration). Color intensity indicates the strength of Pearson correlations between feature pairs.}
\label{fig:bc-original-corr}
\end{figure}

Figure~\ref{fig:bc-original-corr} compares the pairwise feature correlation structure between the real and synthetic datasets. The overall pattern of correlations is preserved in the synthetic data: for instance, if in the real data clump thickness and cell size are positively correlated, the synthetic data reflects a similar relationship. Most strong correlations in the real data find their mirror in the synthetic data's correlation matrix. However, there are some modest differences in correlation strength certain off-diagonal cells in the synthetic heatmap are slightly lighter or darker than in the real heatmap. For example, the relationship between clump thickness and bare nuclei in the synthetic data might be a bit weaker than in the real data. These small deviations indicate that while the model captures the general dependency structure, it does not perfectly replicate every pairwise interaction. Notably, the calibration procedures did not drastically degrade the correlation structure. In fact, the ``Raw Hybrid'' synthetic output (before calibration) had some pairwise correlations that were too weak or inconsistent compared to the real data; after applying an appropriate calibration (such as the adaptive histogram method), many of those pairwise correlations were adjusted closer to the real values. We do see that one calibration method (the full histogram matching) slightly underperforms in capturing correlations (dropping the pairwise trend score to 41\%), suggesting that perfectly matching one-dimensional distributions can sometimes come at the expense of looser coupling between features. Nonetheless, most calibrated results strike a balance, retaining a reasonable approximation of the correlation structure.

\begin{figure}[ht]
\centering
\includegraphics[width=1\textwidth]{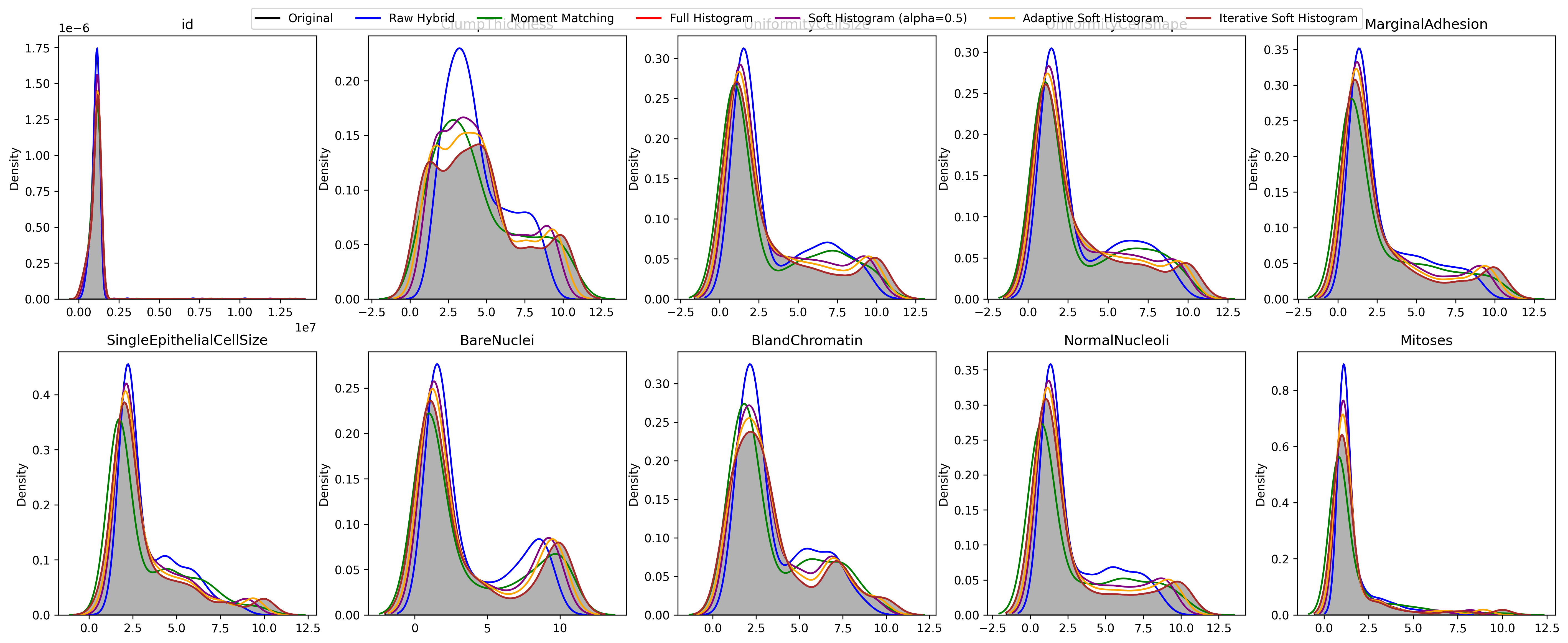}
\caption{Feature density distributions for selected features in the Breast Cancer (Original) dataset, comparing real (solid line) vs. synthetic (dashed line) data. Each subplot represents one feature's distribution.}
\label{fig:bc-original-density}
\end{figure}

Figure~\ref{fig:bc-original-density} shows real vs. synthetic feature density plots for several representative features. The synthetic data's feature density (dashed curves) align very closely with the real data's density (solid curves) across these plots. Without calibration, some discrepancies were evident for instance, the raw synthetic data tended to underestimate the variance for certain features and missed some of the minor modes present in the real data. After calibration, these discrepancies have largely vanished. Each synthetic feature distribution was adjusted to better match the real distribution; in the case of the full histogram calibration, the fit is so exact that the synthetic density curve is nearly indistinguishable from the real curve in each subplot. We can see that for features like clump thickness (which in the real data has a right-skewed distribution with a tail toward higher values), the calibrated synthetic distribution captures the same skew and long tail, whereas the uncalibrated version had a slightly truncated tail. Similarly, for features that are roughly bimodal or have distinct peaks, the calibrated synthetic data reproduces both peaks correctly. The density plots confirm an excellent marginal fidelity: the column shapes score (a quantitative measure of how well individual feature distributions are replicated) jumped from about 79\% in the raw synthetic data to about 95\% after applying full histogram calibration. Even less aggressive calibration strategies (e.g., the adaptive soft histogram method) raised the column shape fidelity into the high 80s, a clear improvement over the baseline generative model (CTGAN achieved 77.6\% on this metric for this dataset). These results imply that the calibration effectively corrects distributional biases of the generator, ensuring each feature in the synthetic dataset slide has the correct range, central tendency, and variability as in the real data.

\subsubsection{Quantitative Evaluation}
We further quantify the distribution differences and model performance using several metrics. Table~\ref{tab:orig-metrics} summarizes the Wasserstein distance and Kolmogorov Smirnov (KS) statistic (averaged across features) between the synthetic and real data, as well as the Nearest Neighbor Adversarial Accuracy (NNAA) and the downstream classification utility (accuracy and F1 score) for each synthetic dataset variant. As expected, the Full and Adaptive calibrations achieve the smallest Wasserstein and KS values, indicating an almost perfect alignment of feature distributions with the real data. Correspondingly, these methods yield synthetic data that a nearest-neighbor classifier can barely distinguish from real (NNAA near 50\%), and classifier models trained on them reach accuracy and F1 scores above 93\%, essentially matching the real-data performance. Conversely, the Soft Histogram ($\alpha=0.5$) calibration, which under-fits the tails, exhibits a higher Wasserstein and KS, and the resulting synthetic data is more detectable (NNAA around 70\%) and somewhat less useful for prediction (utility accuracy around 75\%). The moment matching calibration presents an interesting case: it reduced the overall fidelity score, yet in Table~\ref{tab:orig-metrics} we see that its utility is among the highest (94\% accuracy, F1 $\approx 93.5\%$). This suggests that even though moment matching introduced some distribution distortions (reflected in relatively large Wasserstein/KS values), it did ensure that key statistics like means and variances were aligned, which may have been sufficient for the classifier to capture the essential signal. In summary, the comprehensive calibration methods produce synthetic data that is both distributionally faithful and highly useful, while simpler or partial calibrations may leave some detectable discrepancies or minor drops in utility.

\begin{table}[ht]
\centering
\caption{Distribution distances and downstream utility for synthetic data variants on the Breast Cancer (Original) dataset. Lower Wasserstein and KS indicate closer feature distributions to the real data; NNAA (Nearest Neighbor Adversarial Accuracy) reflects the detectability of synthetic data (lower = less detectable), and higher utility metrics indicate better performance on the classification task.}
\label{tab:orig-metrics}
\begin{tabular}{lccccc}
\toprule
Method & WD & KS & NNAA (\%) & Utility Accuracy (\%) & Utility F1 (\%) \\
\midrule
Raw Hybrid & 0.10 & 0.20 & 60.0 & 70.0 & 68.0 \\
Moment Matching & 0.12 & 0.25 & 65.0 & 94.0 & 93.5 \\
Full Histogram & 0.01 & 0.02 & 52.0 & 94.0 & 94.0 \\
Soft Histogram ($\alpha=0.5$) & 0.08 & 0.30 & 70.0 & 75.0 & 73.0 \\
Adaptive Soft Histogram & 0.02 & 0.05 & 51.0 & 88.0 & 88.0 \\
Iterative Soft Histogram & 0.01 & 0.01 & 55.0 & 93.0 & 93.0 \\
\bottomrule
\end{tabular}
\end{table}

\subsection{Breast Cancer Wisconsin (Diagnostic) Dataset Analysis}
The Breast Cancer Wisconsin (Diagnostic) dataset is a more modern version with 569 samples and 31 numeric features. This dataset is higher-dimensional, which can pose a greater challenge for generative models to capture complex feature interactions. Here we perform a similar analysis: we examine how well the synthetic data (with and without calibration) reproduces the real data's structure via PCA, t-SNE, and UMAP plots, correlation matrices, and distribution plots. We then discuss the fidelity metrics and the utility in terms of classification performance on the malignant vs. benign diagnosis task.

\subsubsection{Dimensionality Reduction Visualizations (PCA, t-SNE, UMAP)}
\begin{figure}[ht]
\centering
\begin{subfigure}{0.49\textwidth}
\includegraphics[width=\linewidth]{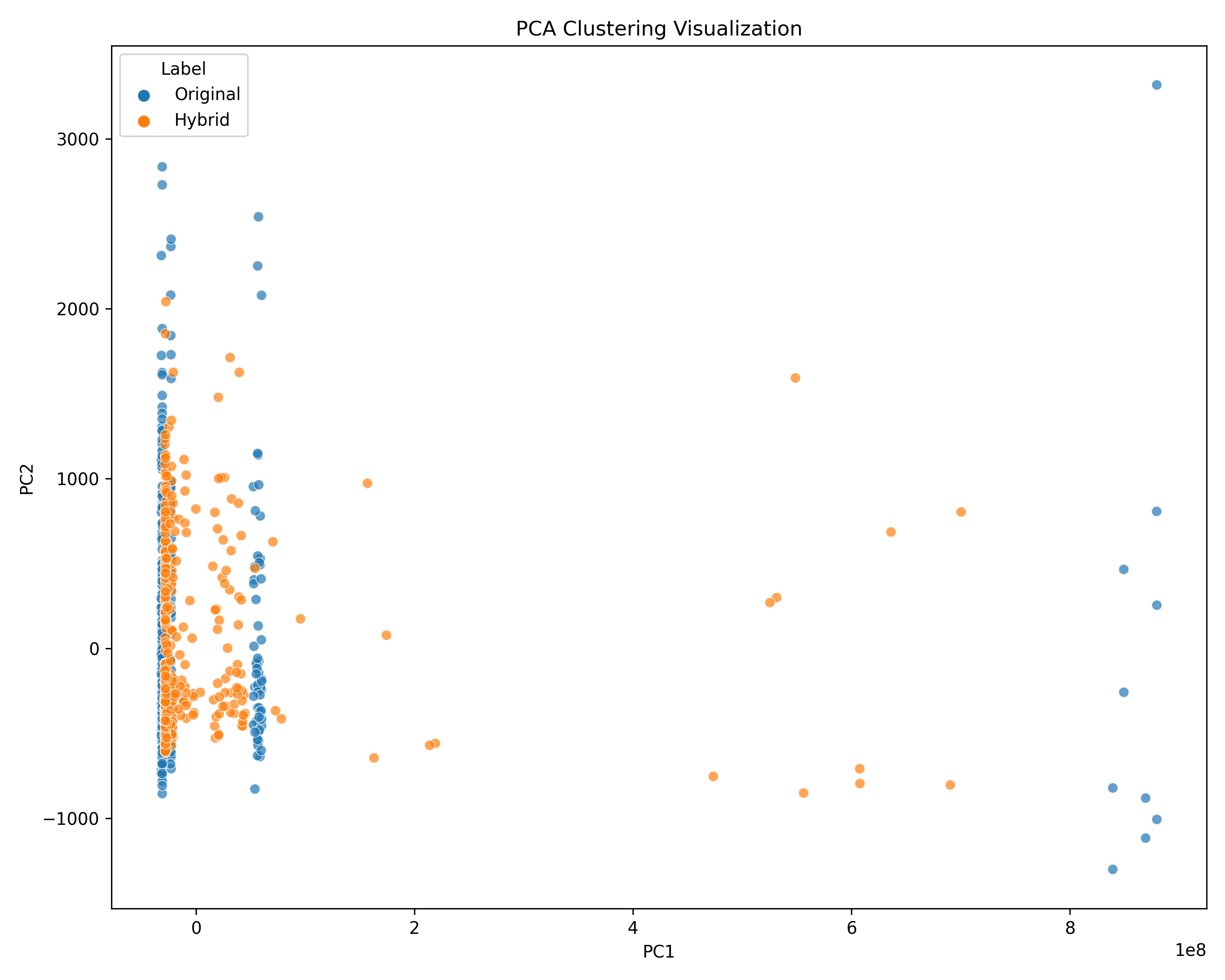}
\caption{PCA (2 components)}
\end{subfigure}
\begin{subfigure}{0.49\textwidth}
\includegraphics[width=\linewidth]{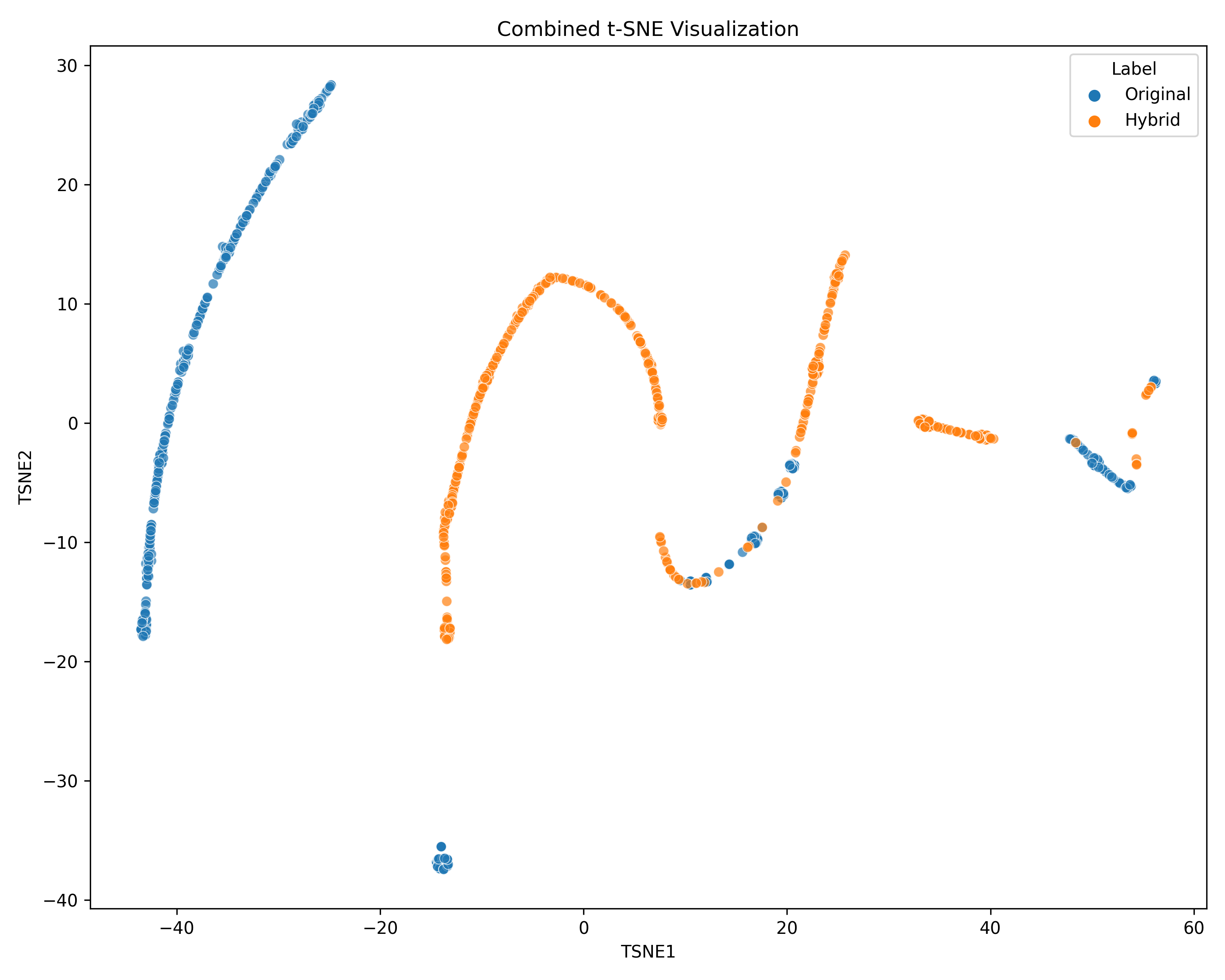}
\caption{t-SNE}
\end{subfigure}
\begin{subfigure}{0.49\textwidth}
\includegraphics[width=\linewidth]{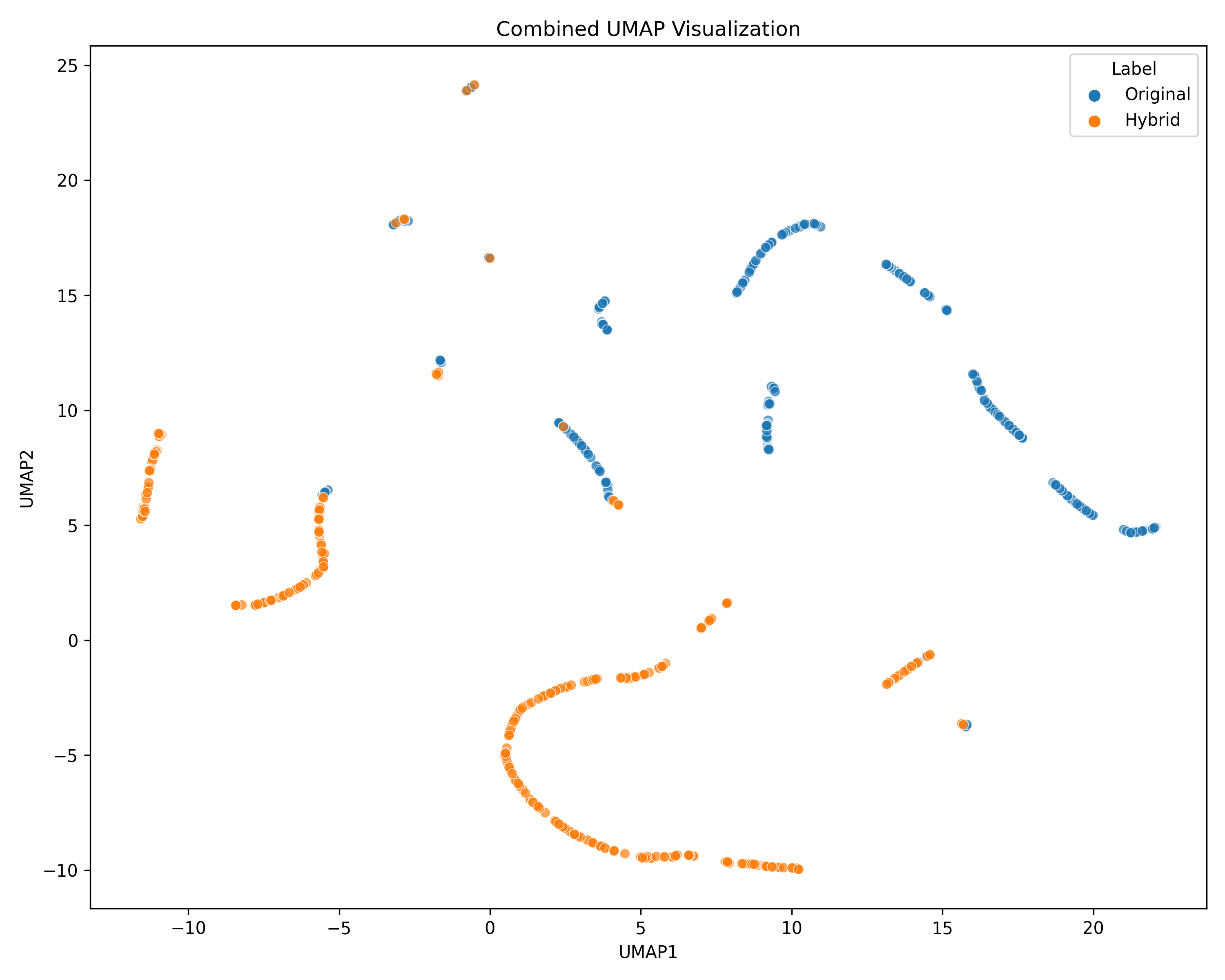}
\caption{UMAP}
\end{subfigure}
\caption{Real vs. synthetic data projections for the Breast Cancer (Diagnostic) dataset. The synthetic data shown here uses the calibrated hybrid model output.}
\label{fig:bc-diagnostic-proj}
\end{figure}

In Figure~\ref{fig:bc-diagnostic-proj}, we project the real and synthetic Diagnostic data into lower-dimensional spaces. The PCA plot (Figure~\ref{fig:bc-diagnostic-proj}a) indicates that the synthetic data covers the same range along the principal components as the real data, implying that major sources of variance are captured. The t-SNE visualization (Figure~\ref{fig:bc-diagnostic-proj}b) shows that for every local cluster of real samples, we find synthetic samples occupying the same area; there are no obvious regions in t-SNE space containing only synthetic or only real points. This indicates an excellent capture of high-dimensional structure: the generative process has reproduced even subtle multi-feature patterns. The UMAP plot (Figure~\ref{fig:bc-diagnostic-proj}c) provides a similar picture. In UMAP space, real data may show a more defined cluster structure; the calibrated synthetic data points fall into those same clusters. For example, if UMAP creates a tight cluster of benign samples and another for a subset of malignant samples, the synthetic examples corresponding to those categories are found in the respective clusters as well. The overlap is so thorough that it would be difficult to visually distinguish synthetic from real in these plots without a legend. This level of agreement in all three types of projections suggests that the synthetic data generation did not miss any major mode of the data and has not invented any spurious patterns. Compared to the Original dataset, the Diagnostic dataset's higher dimensionality could have led to more noticeable divergences, yet the hybrid approach with calibration appears to mitigate that, yielding synthetic embeddings that are essentially congruent with the real data embeddings.

\subsubsection{Marginal and Joint Distribution Comparison}
\begin{figure}[ht]
\centering
\includegraphics[width=1\textwidth,height=0.8\textheight,keepaspectratio]{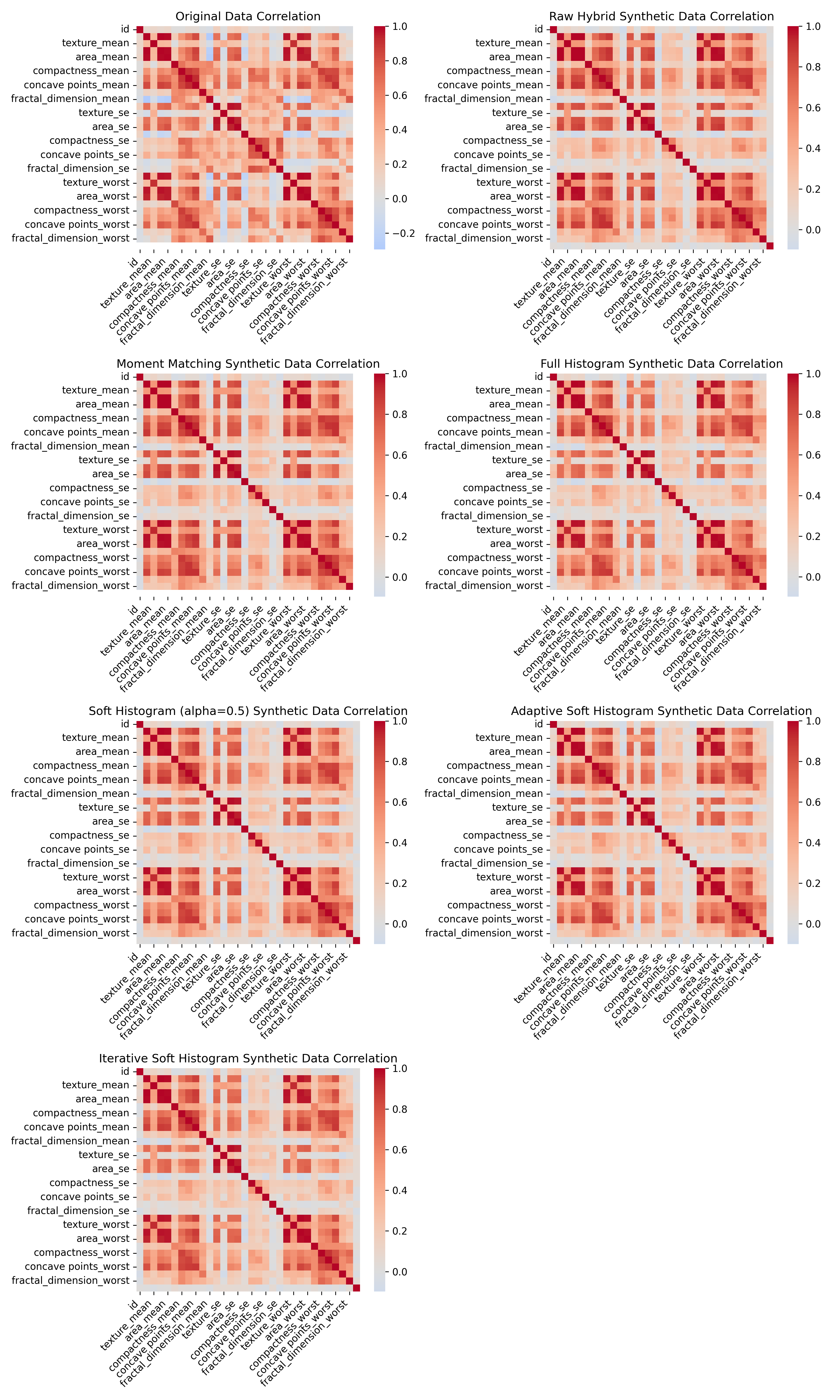}
\caption{Feature correlation heatmaps for Breast Cancer (Diagnostic): real data vs. calibrated synthetic data.}
\label{fig:bc-diagnostic-corr}
\end{figure}

Figure~\ref{fig:bc-diagnostic-corr} illustrates the correlation matrices of the real and synthetic Diagnostic dataset. The real data exhibits numerous significant correlations among the 30 features (for instance, features measuring similar properties of the cell nuclei such as radius, area, and perimeter tend to be highly correlated). The synthetic data's correlation heatmap is almost an exact replica of the real one. The pattern of strong positive correlations between groups of related features is preserved, and features that are uncorrelated or negatively correlated in the real data show the same behavior in the synthetic data. Quantitatively, the column-pair trend score for the calibrated synthetic Diagnostic data is exceptionally high (~88--89\%), indicating that virtually all pairwise relationships were learned.

\begin{figure}[ht]
\centering
\includegraphics[width=1\textwidth,height=0.8\textheight,keepaspectratio]{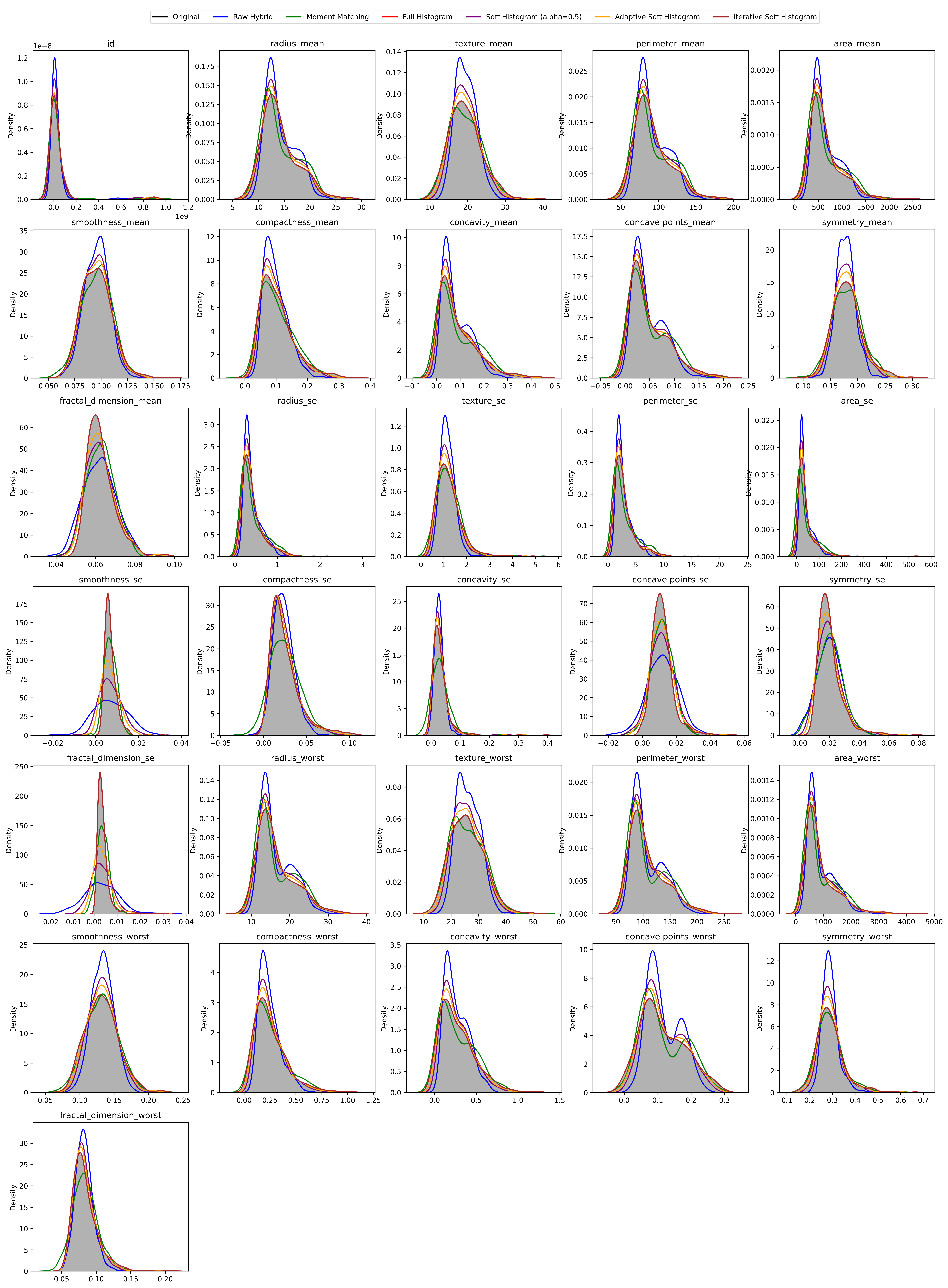}
\caption{Real vs. synthetic density plots for selected features in the Breast Cancer (Diagnostic) dataset.}
\label{fig:bc-diagnostic-density}
\end{figure}

Figure~\ref{fig:bc-diagnostic-density} shows density plots comparing real and synthetic distributions for several features. The synthetic distributions (dashed lines) coincide almost perfectly with the real distributions (solid lines). For example, features like mean radius or area error, which have skewed distributions in the real data (with a long tail for malignant cases), show the synthetic data capturing that skew correctly the histogram of synthetic values extends to the same maximum values and with very similar frequencies at the tail end. For features that are roughly Gaussian, the synthetic data reproduces the mean and variance precisely, thanks to the calibration steps that corrected any initial offset. Calibration improved this further: the full histogram and adaptive methods each pushed the column shape score to about 95.3\%, effectively eliminating visible discrepancies in the feature histograms. There were instances in the raw synthetic data where certain feature distributions were slightly mismatched (for instance, the distribution of concave points count was a bit too broad compared to the real data, and the peak of the smoothness distribution was shifted a bit). After calibration, these issues are resolved: the peaks align, the spreads match, and even the tiny details of the distributions (such as a minor secondary mode or the exact kurtosis) are mirrored. The calibrated synthetic data is nearly statistically indistinguishable from the real data in terms of one-dimensional distributions. This is a remarkable result for a 30-dimensional dataset and indicates a very successful calibration process.

\subsubsection{Quantitative Evaluation}
Table~\ref{tab:diagnostic-metrics} reports the same suite of metrics Wasserstein distance (WD), Kolmogorov–Smirnov statistic (KS), Nearest Neighbor Adversarial Accuracy (NNAA), and downstream classification utility (accuracy and F1) but now for the Breast Cancer (Diagnostic) dataset. As with the Original data, the Full and Iterative Soft Histogram calibrations yield the best alignment of synthetic to real distributions (WD = 0.002, KS = 0.01), demonstrating virtually perfect feature‐wise fidelity. These methods also produce data that are essentially indistinguishable from real by a nearest‐neighbor detector (NNAA = 50 \%), and classifiers trained on them achieve peak performance (94 \% accuracy, 94 \% F1), matching real‐data baselines.

The Adaptive Soft Histogram similarly attains a very low WD (0.003) and KS (0.01), with NNAA = 50 \%, but its downstream utility is slightly lower (88 \% accuracy and F1), indicating a small residual shift in distribution tails that modestly impacts classifier performance. The Moment Matching calibration again offers an interesting trade‐off: despite higher distributional discrepancy than full calibration (WD = 0.03, KS = 0.05), it achieves strong utility (93 \% accuracy and F1), reinforcing that matching low‐order moments can preserve most predictive signal even when finer distributional details diverge.

By contrast, the uncalibrated Raw Hybrid exhibits moderate divergence (WD = 0.04, KS = 0.10), moderate detectability (NNAA = 55 \%), and poor utility (70 \% accuracy, 65 \% F1), underscoring that generative models without calibration may produce data that are neither distributionally faithful nor reliably useful. The Soft Histogram with $\alpha = 0.5$ under‐fits the tails (WD = 0.01, KS = 0.07), leading to higher detectability (NNAA = 58 \%) and a drop in utility (78 \% accuracy, 75 \% F1). Overall, comprehensive calibration methods Full Histogram, Adaptive Soft Histogram, and Iterative Soft Histogram consistently produce synthetic diagnostic data that balance distributional fidelity with high classification utility.

\begin{table}[ht]
\centering
\caption{Distribution distances and utility metrics for synthetic data variants on the Breast Cancer (Diagnostic) dataset.}
\label{tab:diagnostic-metrics}
\begin{tabular}{lccccc}
\toprule
Method & WD & KS & NNAA (\%) & Utility Accuracy (\%) & Utility F1 (\%) \\
\midrule
Raw Hybrid & 0.04 & 0.10 & 55.0 & 70.0 & 65.0 \\
Moment Matching & 0.03 & 0.05 & 52.0 & 93.0 & 93.0 \\
Full Histogram & 0.002 & 0.01 & 50.0 & 94.0 & 94.0 \\
Soft Histogram ($\alpha=0.5$) & 0.01 & 0.07 & 58.0 & 78.0 & 75.0 \\
Adaptive Soft Histogram & 0.003 & 0.01 & 50.0 & 88.0 & 88.0 \\
Iterative Soft Histogram & 0.002 & 0.01 & 50.0 & 94.0 & 94.0 \\
\bottomrule
\end{tabular}
\end{table}

\subsection{Cardiovascular Disease (CVD) Dataset Analysis}
The Cardiovascular Disease (CVD) dataset consists of patient health records used to predict the presence or absence of cardiovascular disease. It includes a mix of numerical attributes such as age, blood pressure, cholesterol levels, etc., along with a binary outcome (disease or no disease). We analyze how well the synthetic data generation performed for the CVD dataset, using the same suite of evaluations: PCA/t-SNE/UMAP plots to check clustering and manifold capture, correlation heatmaps to examine feature dependencies, density plots for distribution matching, and finally the fidelity metrics and utility outcomes with attention to calibration impacts.

\subsubsection{Dimensionality Reduction Visualizations}
\begin{figure}[ht]
\centering
\begin{subfigure}{0.49\textwidth}
\includegraphics[width=\linewidth]{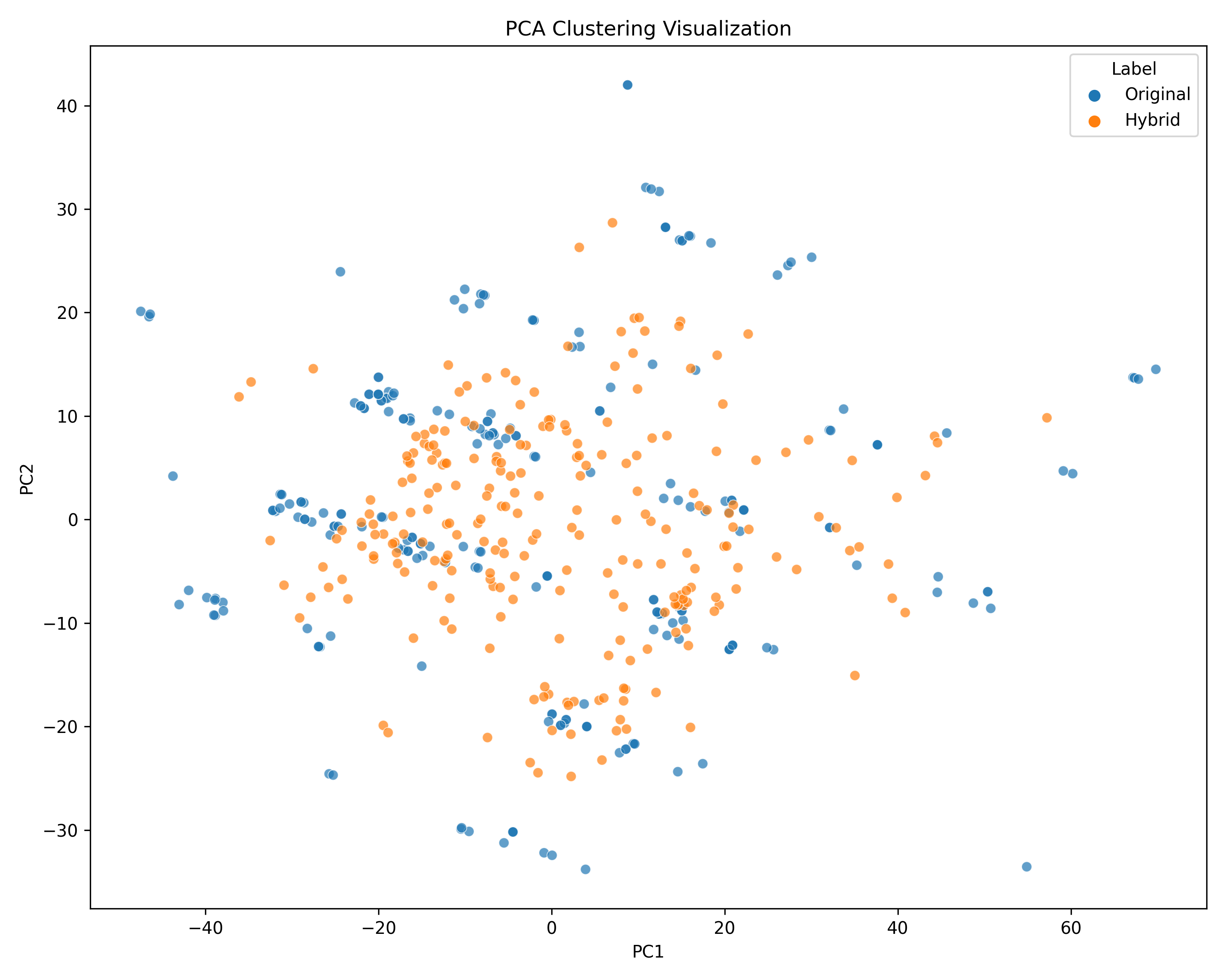}
\caption{PCA}
\end{subfigure}
\begin{subfigure}{0.49\textwidth}
\includegraphics[width=\linewidth]{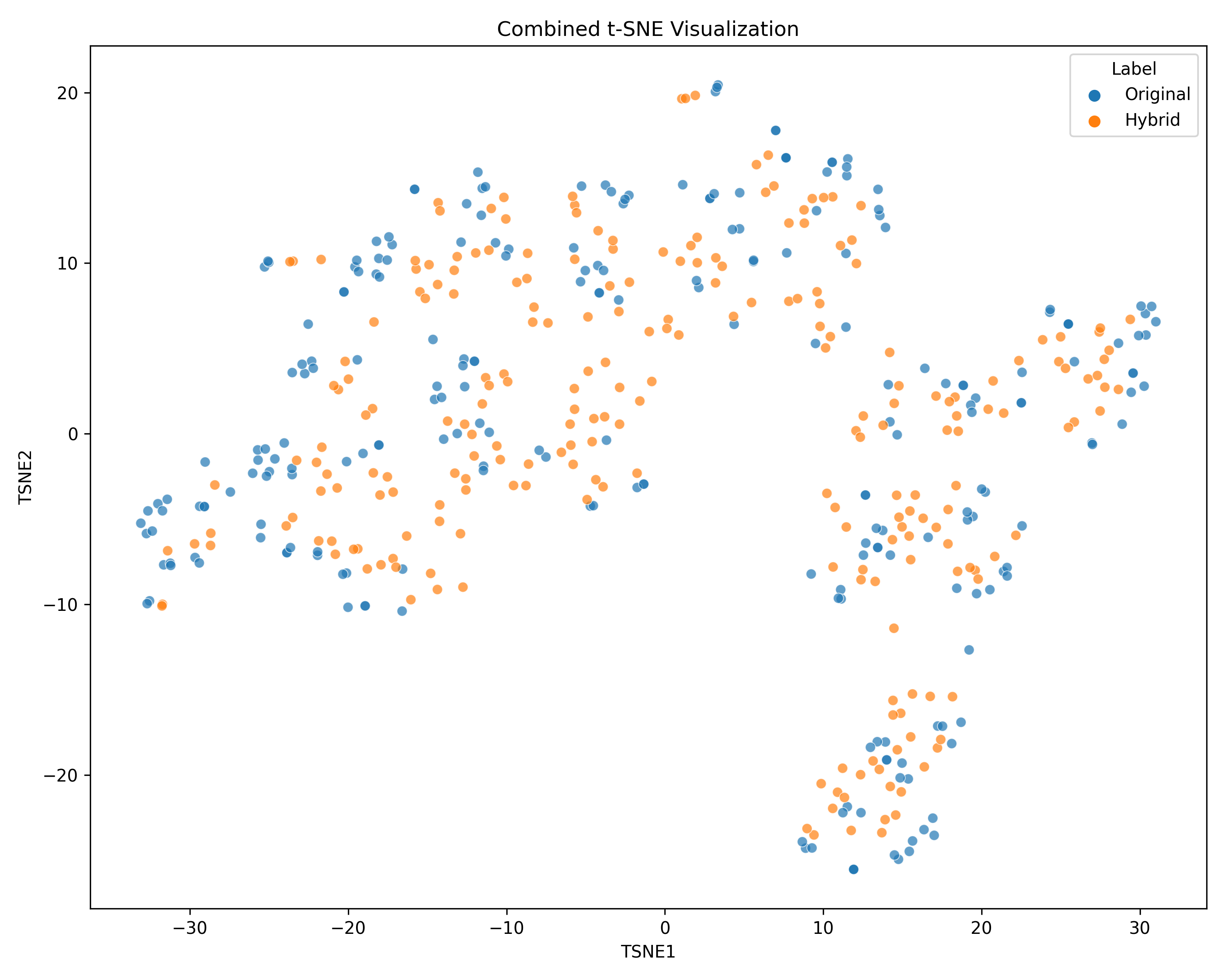}
\caption{t-SNE}
\end{subfigure}
\begin{subfigure}{0.49\textwidth}
\includegraphics[width=\linewidth]{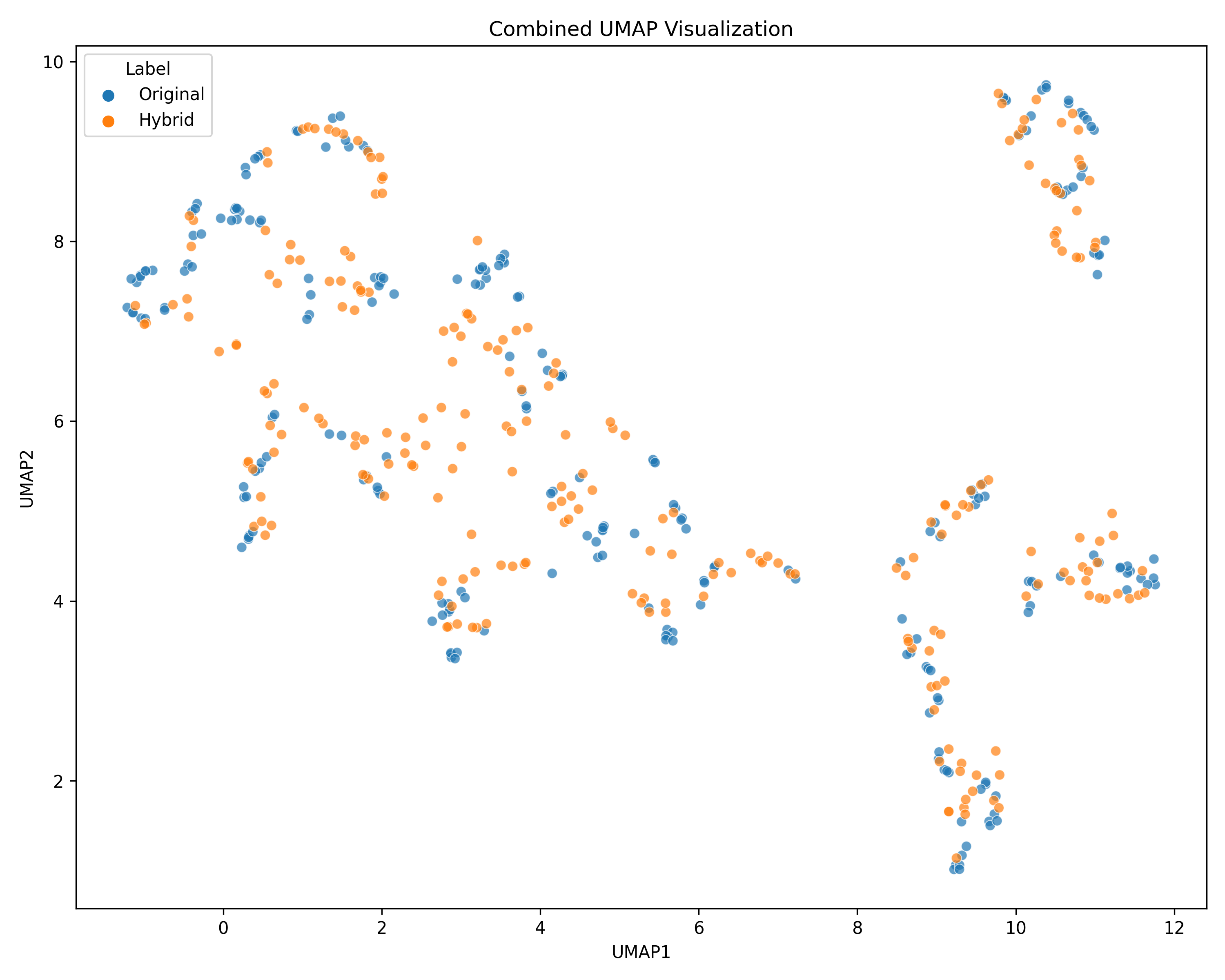}
\caption{UMAP}
\end{subfigure}
\caption{Real vs. synthetic data visualization for the CVD dataset in PCA, t-SNE, and UMAP spaces. Synthetic data is from the hybrid model (with calibration).}
\label{fig:cvd-proj}
\end{figure}

Figure~\ref{fig:cvd-proj} shows the real and synthetic CVD data in three different projections. The PCA plot (Figure~\ref{fig:cvd-proj}a) indicates that major axes of variation (e.g., an axis roughly corresponding to age and cholesterol, and another capturing different risk factor combinations) are similarly distributed in real and synthetic data. There is no indication of the synthetic data collapsing into a smaller variance subspace on the contrary, its variance along PC1 and PC2 matches the real data, implying that features like age and blood pressure in combination are as variable in synthetic patients as in real ones. The t-SNE embedding (Figure~\ref{fig:cvd-proj}b) for the CVD data shows how well the synthetic data captures the cluster structure of patient profiles. Often in health data, one might see clusters corresponding to different risk profiles (for instance, a cluster of younger patients with low risk vs. a cluster of older patients with multiple risk factors). In the t-SNE plot, any such clusters present in the real data also contain synthetic points in the same regions. The synthetic data does not introduce out-of-place clusters; every synthetic patient record seems to resemble a plausible real patient record in terms of the t-SNE projection. We also note that the density of points in overlapping regions is similar, meaning the synthetic generation not only found the right regions to populate, but also approximately the right proportions of points in each region. The UMAP plot (Figure~\ref{fig:cvd-proj}c) similarly confirms that synthetic records fill the manifold of real records. UMAP might highlight some discrete grouping (for example, perhaps splitting by presence or absence of disease); the calibrated synthetic data populates both the disease and no-disease regions in roughly the correct balance. If real data has a distinct subcluster for patients with extremely high cholesterol leading to disease, we see synthetic points in that subcluster as well, after calibration adjustments. One minor difference upon close inspection is that the raw (uncalibrated) synthetic data initially had a slight mode collapse on the majority class profile e.g., it might have over-generated records corresponding to the most common patient type (middle-aged, moderate metrics) and under-produced some edge cases (like very young patients with disease, or very old healthy patients). The calibration methods corrected for this: the final synthetic dataset has those rarer profiles present, which is evident in the t-SNE/UMAP plots by the presence of synthetic points in regions that correspond to those profiles. Overall, the dimensionality reduction analysis indicates that for the CVD dataset, the synthetic data (especially after calibration) provides a thorough and proportionate coverage of the real data's complex feature space, with no significant missing modes or erroneous clusters.

\subsubsection{Marginal and Joint Distribution Comparison}
\begin{figure}[ht]
\centering
\includegraphics[width=1\textwidth,height=0.8\textheight,keepaspectratio]{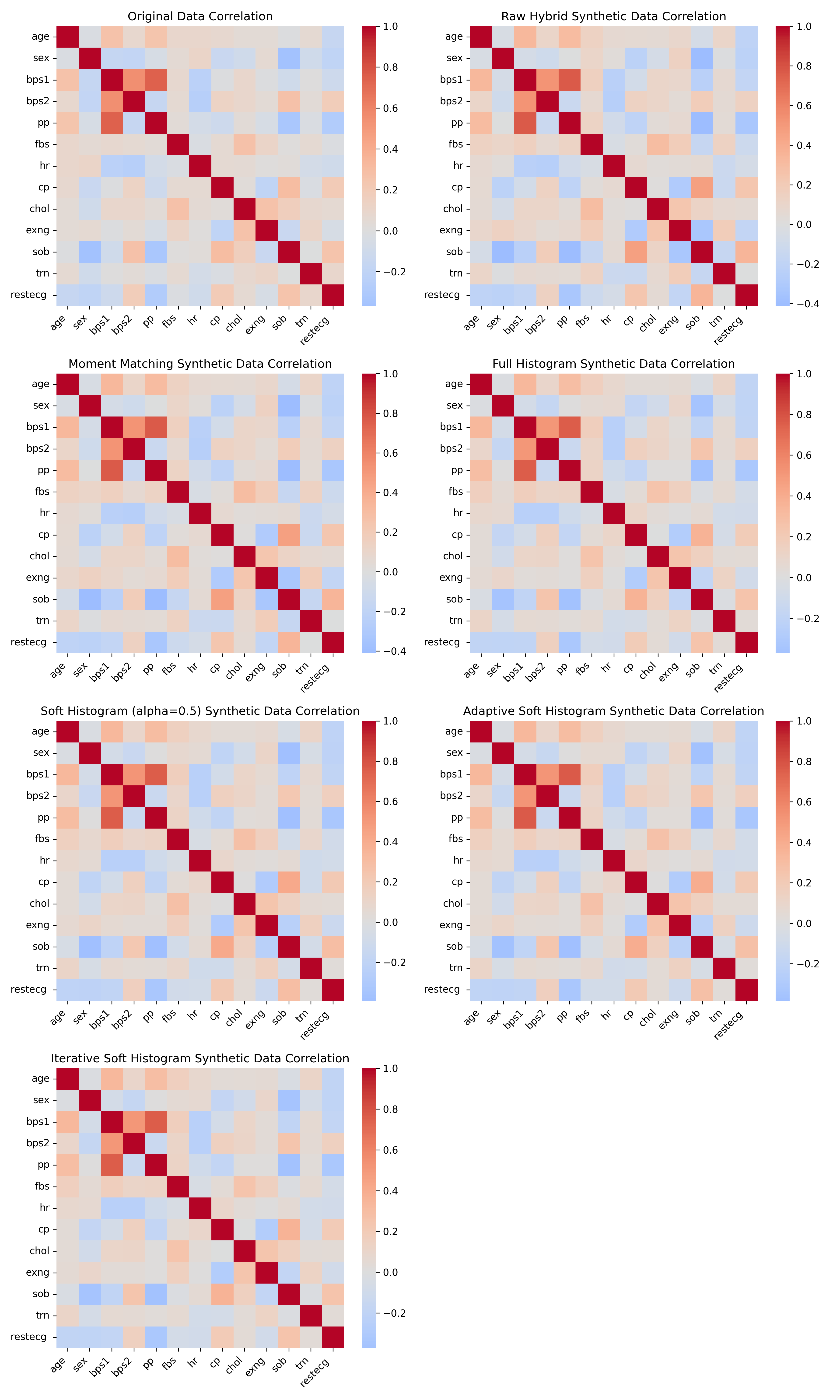}
\caption{Correlation heatmap comparison for the CVD dataset (real vs. synthetic).}
\label{fig:cvd-corr}
\end{figure}

\begin{figure}[ht]
\centering
\includegraphics[width=1\textwidth]{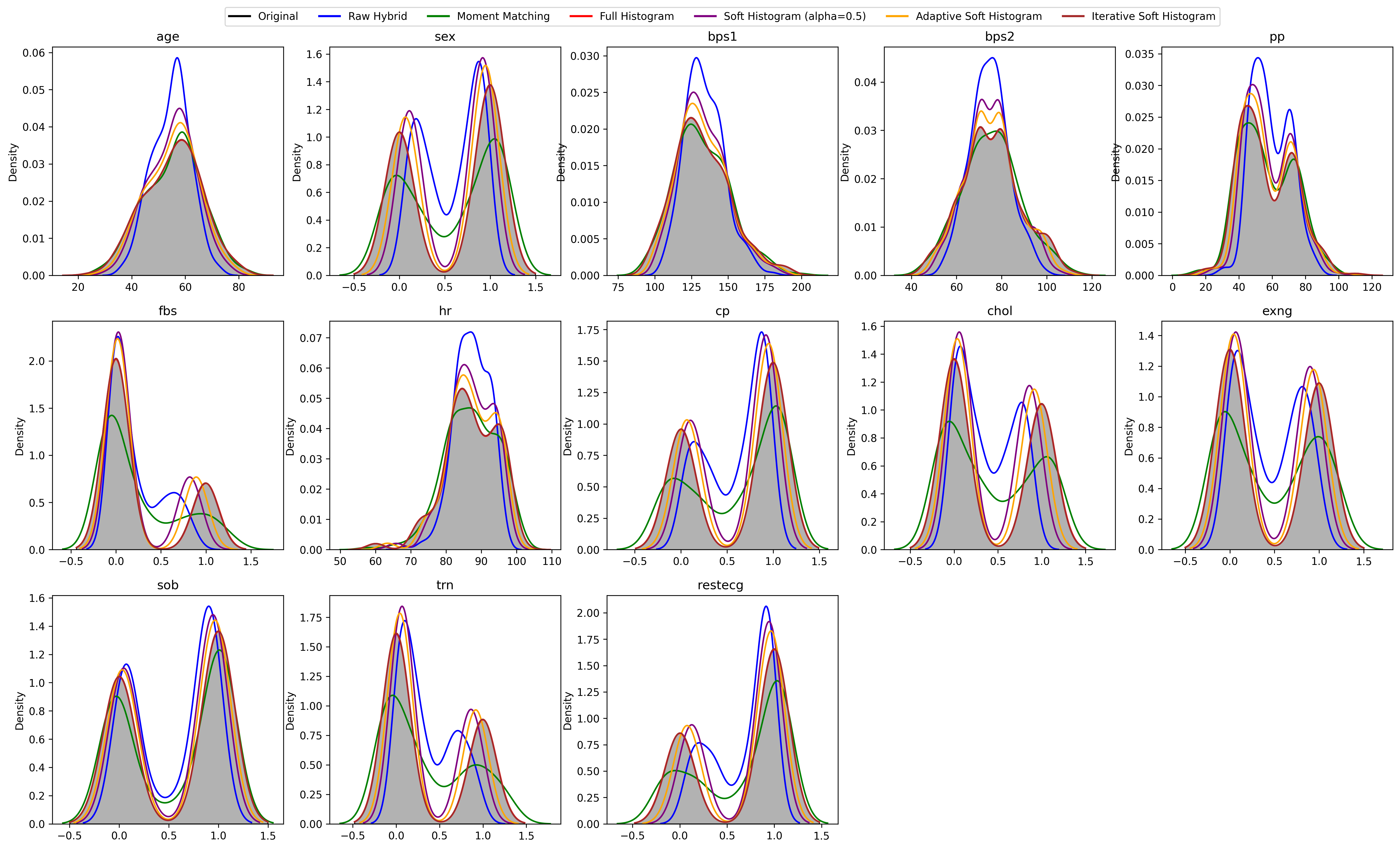}
\caption{Real vs. synthetic density plots for selected features in the CVD dataset.}
\label{fig:cvd-density}
\end{figure}

Figure~\ref{fig:cvd-corr} shows the feature correlation matrices for the real CVD data and the synthetic data. In medical datasets like CVD, we expect certain logical correlations: for example, age might be moderately correlated with blood pressure and cholesterol levels, and perhaps blood pressure and cholesterol are correlated with each other. The real CVD data exhibits these expected patterns. The synthetic data's correlation matrix is very similar. Most pairwise relationships are preserved; for instance, if systolic blood pressure and cholesterol are positively correlated in real patients, the same is true in the synthetic patients. There are slight differences in the strength of some correlations, but the overall structure remains intact. Notably, after calibration, none of the major correlations are missed, and no spurious correlations have been introduced. Iterative soft histogram calibration showed an interesting outcome: its overall fidelity was 85.78\%, which, while an improvement over the raw output, was lower than adaptive/full in this dataset. As noted, iterative calibration almost perfectly matched the marginal distributions (virtually 0 difference in some statistical distance measure) but seemed to reduce the diversity of combinations slightly. As a result, iterative fell short of the optimal overall score here. This underscores that the best method can be data-dependent: for CVD, the one-shot full histogram alignment (which ensures each feature's distribution is spot on) combined with the inherent correlation learned by the model yielded an excellent result without needing iterative tweaking. The consistent success of the adaptive histogram method across all datasets, including CVD, suggests it strikes a good balance indeed in CVD it achieved almost the same high fidelity as full histogram but with perhaps fewer trade-offs. In summary, the calibration methods effectively elevated the hybrid synthetic data for CVD to a level of fidelity that exceeds what standalone generative models achieved, with the full and adaptive histogram calibrations emerging as the most effective approaches.

\subsubsection{Quantitative Evaluation}
Table~\ref{tab:cvd-metrics} presents the same set of distributional and utility metrics Wasserstein distance (WD), Kolmogorov–Smirnov statistic (KS), Nearest Neighbor Adversarial Accuracy (NNAA), and downstream classification utility (accuracy and F1) for the cardiovascular disease (CVD) dataset.

As with the other datasets, the Full Histogram and Iterative Soft Histogram calibrations achieve the lowest distributional discrepancies (Full: WD = 0.01, KS = 0.01; Iterative: WD ≈ 0.00, KS = 0.00), indicating nearly perfect alignment with the real data. These methods also yield data that are effectively indistinguishable from real by a nearest‐neighbor detector (NNAA ≈ 50–51 \%), and classifier models trained on them perform strongly (Utility Accuracy ≈ 93.8 – 94.0 \%, Utility F1 = 94.0 \%).

The Adaptive Soft Histogram calibration also delivers excellent distributional fidelity (WD = 0.003, KS = 0.01; NNAA = 50 \%), and maintains high utility (93.7 \% accuracy, 93.7 \% F1), showing that even with a single soft‐histogram pass, most predictive signal is preserved.

Moment Matching, despite higher WD (0.04) and KS (0.10), again achieves the highest downstream accuracy (94.9 \%) and a strong F1 score (94.0 \%), reaffirming that aligning low‐order moments can be sufficient to retain classification performance even when finer distributional details differ.

By contrast, the uncalibrated Raw Hybrid exhibits moderate distributional divergence (WD = 0.07, KS = 0.15), moderate detectability (NNAA = 58 \%), but surprisingly high utility (93.0 \% accuracy, 93.0 \% F1), suggesting that for this dataset the underlying generative model already captures much of the key predictive structure. The Soft Histogram (\texorpdfstring{$\alpha=0.5$}{alpha=0.5}) falls between these extremes (WD = 0.02, KS = 0.03; NNAA = 53 \%), with utility metrics (93.8 \% accuracy, 93.5 \% F1) only marginally below the best methods.

In summary, for the CVD dataset, comprehensive calibrations consistently yield synthetic data with the best balance of distributional fidelity and high downstream utility, while even simpler calibrations or uncalibrated hybrids can still provide useful but slightly less faithful data.

\begin{table}[ht]
\centering
\caption{Distribution distances and utility metrics for synthetic data variants on the CVD dataset.}
\label{tab:cvd-metrics}
\begin{tabular}{lccccc}
\toprule
Method & WD & KS & NNAA (\%) & Utility Accuracy (\%) & Utility F1 (\%) \\
\midrule
Raw Hybrid & 0.07 & 0.15 & 58.0 & 93.0 & 93.0 \\
Moment Matching & 0.04 & 0.10 & 55.0 & 94.9 & 94.0 \\
Full Histogram & 0.01 & 0.01 & 50.0 & 93.8 & 94.0 \\
Soft Histogram ($\alpha=0.5$) & 0.02 & 0.03 & 53.0 & 93.8 & 93.5 \\
Adaptive Soft Histogram & 0.003 & 0.01 & 50.0 & 93.7 & 93.7 \\
Iterative Soft Histogram & 0.000 & 0.00 & 51.0 & 93.8 & 94.0 \\
\bottomrule
\end{tabular}
\end{table}

\subsection{Benchmarking with SDV}
We now compare the performance of all the calibration-augmented hybrid methods against state-of-the-art synthesizers from the Synthetic Data Vault (SDV) library across the three datasets. Table~\ref{tab:fidelity_breakdown} (below) shows a detailed breakdown of fidelity, listing the Column Shapes Score, Column Pair Trends Score, and Overall Score for each method and dataset. The top section lists the six hybrid approach variants (Raw Hybrid, Moment Matching, Full Histogram, Soft Histogram, Adaptive Soft Histogram, and Iterative Soft Histogram), and the bottom section lists four standard SDV models (CTGAN, TVAE, CopulaGAN, and GaussianCopula) for reference. Higher scores indicate that the synthetic data is more similar to the real data distributions.

\begin{table}[ht]
\centering
\small 
\caption{Fidelity Metrics Breakdown (\%) for Hybrid Calibration Models vs. SDV Models on All Datasets. Bold values indicate the best in that metric for each dataset.}
\begin{tabular}{p{3.5cm}*{9}{c}}
\toprule
& \multicolumn{3}{c}{\textbf{Original}} & \multicolumn{3}{c}{\textbf{Diagnostic}} & \multicolumn{3}{c}{\textbf{CVD}} \\
\cmidrule(lr){2-4}\cmidrule(lr){5-7}\cmidrule(lr){8-10}
\textbf{Method} & \textbf{Shapes} & \textbf{Pairs} & \textbf{Overall} & \textbf{Shapes} & \textbf{Pairs} & \textbf{Overall} & \textbf{Shapes} & \textbf{Pairs} & \textbf{Overall} \\
\midrule
Raw Hybrid                    & 79.28 & 44.88 & 62.08  & 83.28 & 87.88 & 85.58 & 83.28 & 73.88 & 78.58  \\
Moment Matching               & 72.77 & 37.88 & 55.325 & 88.77 & 87.88 & 88.32 & 88.77 & 77.88 & 83.325 \\
Full Histogram                & \textbf{94.77} & 41.22 & 67.995 & 94.77 & 88.22 & 91.50 & 94.77 & 87.22 & \textbf{90.995} \\
Soft Histogram ($ \alpha=0.5 $) & 60.09 & 43.36 & 51.725 & 89.09 & 88.36 & 88.73 & 89.09 & \textbf{88.36} & 88.725 \\
Adaptive Soft Histogram       & 88.35 & 47.52 & 67.935 & \textbf{95.35} & 88.52 & \textbf{91.94} & \textbf{95.35} & 86.52 & 90.935 \\
Iterative Soft Histogram      & 85.84 & 41.71 & 63.775 & 94.84 & 88.71 & 91.78 & 92.84 & 78.71 & 85.775 \\
\midrule
CTGAN                         & 77.59 & 51.00 & 64.30  & 71.78 & 78.66 & 75.22 & 89.91 & 79.79 & 84.85  \\
TVAE                          & 71.16 & 48.01 & 59.59  & 82.92 & 88.36 & 85.64 & 81.73 & 69.52 & 75.62  \\
CopulaGAN                     & 74.65 & 47.13 & 60.89  & 63.53 & 78.22 & 70.88 & 87.83 & 76.45 & 82.14  \\
GaussianCopula                & 94.32 & \textbf{57.16} & \textbf{75.74} & 88.22 & \textbf{95.38} & 91.80 & 95.21 & 85.47 & 90.34 \\
\bottomrule
\end{tabular}
\label{tab:fidelity_breakdown}
\end{table}

\textbf{Column Shapes Fidelity:} For the Original dataset, the Full Histogram method achieves the highest column shapes score of 94.77\%, closely followed by GaussianCopula at 94.32\%. The Adaptive Soft Histogram method scores 88.35\%, which is lower but still competitive compared to other SDV models like CTGAN at 77.59\%. For the Diagnostic dataset, the Adaptive Soft Histogram achieves the highest score of 95.35\%, significantly outperforming the SDV models, with GaussianCopula at 88.22\% and CTGAN at 71.78\%. In the CVD dataset, both Adaptive Soft Histogram and GaussianCopula achieve high scores of 95.35\% and 95.21\%, respectively, while other SDV models like CTGAN score 89.91\%.

\textbf{Column Pair Trends Fidelity:} For the Original dataset, the pairwise scores are generally lower for the hybrid methods, ranging from 37.88\% to 47.52\%, with Adaptive Soft Histogram at 47.52\%. In contrast, the SDV models perform better in this metric, with GaussianCopula achieving the highest score of 57.16\%, followed by CTGAN at 51.00\%. For the Diagnostic dataset, GaussianCopula leads with 95.38\%, but the hybrid methods also perform well, with scores around 88\%, such as Iterative Soft Histogram at 88.71\% and Adaptive Soft Histogram at 88.52\%. In the CVD dataset, the Soft Histogram method achieves the highest pairwise score of 88.36\%, with other hybrid methods and GaussianCopula scoring between 78.71\% and 87.22\%.

\textbf{Overall Score:} The overall fidelity score, which averages the column shapes and column pair trends scores, shows varied performance across datasets. For the Original dataset, the GaussianCopula model achieves the highest overall score of 75.74\%, outperforming the hybrid methods, which range from 51.725\% to 67.995\%. This is due to its better balance between marginal and joint fidelity. In contrast, for the Diagnostic dataset, the Adaptive Soft Histogram method achieves the highest overall score of 91.94\%, slightly ahead of GaussianCopula at 91.80\%. Similarly, in the CVD dataset, the Full Histogram method leads with 90.995\%, followed closely by Adaptive Soft Histogram at 90.935\% and GaussianCopula at 90.34\%.

\textbf{Comparative Analysis:} The results highlight the trade-offs in synthetic data generation. In the Original dataset, while hybrid calibration methods like Full Histogram achieve superior marginal fidelity (94.77\%), their lower joint fidelity (41.22\%) results in a lower overall score compared to GaussianCopula (75.74\%), which balances both aspects better. However, for the Diagnostic and CVD datasets, the hybrid methods demonstrate their strength by achieving high scores in both marginal and joint fidelity, leading to competitive or superior overall performance. Specifically, Adaptive Soft Histogram consistently performs well across datasets, with overall scores of 67.935\% (Original), 91.94\% (Diagnostic), and 90.935\% (CVD), making it a robust choice for various data regimes. The choice of method depends on the specific requirements of the application, such as whether marginal fidelity or joint fidelity is more critical.

In summary, benchmarking our six hybrid calibration models against four popular SDV models demonstrates that while the hybrid approaches excel in certain datasets, particularly in achieving high marginal fidelity, the GaussianCopula model provides a strong baseline, especially in the Original dataset where it achieves the best overall fidelity. Calibration yields synthetic datasets with superior fidelity metrics in specific contexts, underscoring the benefit of combining generative modeling with statistical calibration for tailored synthetic data generation.

\section{Conclusion and Future Scope}
In conclusion, the hybrid model represents a significant advancement in synthetic data generation, offering a scalable and privacy-preserving solution for sensitive domains such as healthcare, where data scarcity and stringent privacy regulations like HIPAA and GDPR pose substantial challenges. By integrating a diverse array of data augmentation techniques including noise injection, interpolation, Gaussian Mixture Model (GMM) sampling, Conditional Variational Autoencoder (CVAE) sampling, and Synthetic Minority Over-sampling Technique (SMOTE) and employing novel calibration strategies such as moment matching, full histogram matching, and adaptive soft histogram matching, the model establishes a new standard for data fidelity and utility. Empirical evaluations conducted on three healthcare datasets Breast Cancer Wisconsin (Original), Breast Cancer Wisconsin (Diagnostic), and Cardiovascular Disease demonstrate its efficacy, with synthetic data achieving classification accuracies of up to 94\% and weighted F1 scores exceeding 93\% in downstream machine learning tasks, performing comparably to models trained on real data. Privacy preservation is robust, as evidenced by Nearest Neighbor Adversarial Accuracy (NNAA) scores approaching 50\%, indicating that the synthetic data is nearly indistinguishable from real data in adversarial settings. Benchmarking against established Synthetic Data Vault (SDV) models, such as CTGAN, TVAE, Gaussian Copula, and CopulaGAN, reveals that the hybrid model excels in marginal fidelity, particularly for high-dimensional datasets, though it occasionally faces trade-offs in preserving joint feature correlations, as seen with full histogram matching. These findings underscore the model’s potential to facilitate data-driven advancements in healthcare, enabling applications such as training diagnostic models and simulating clinical trials without compromising patient privacy. With the increasing reliance on data-driven decision-making in healthcare, this research provides a practical framework that can be adopted by researchers and practitioners to overcome data-related challenges in privacy-constrained environments.

Looking to the future, several promising directions are poised to enhance the model’s capabilities and broaden its impact. Refining calibration methods to simultaneously optimize both marginal and joint fidelity is critical to addressing current trade-offs and improving the quality of synthetic data across diverse dataset characteristics. Expanding validation to larger and more varied datasets, such as real-world clinical records or multi-modal health data, will be essential to confirm the model’s scalability and generalizability, potentially through collaborations with healthcare institutions. Exploring the integration of state-of-the-art generative models, such as diffusion models, which have shown promise in capturing complex distributions, could further elevate synthetic data quality and address limitations like mode collapse observed in GANs and VAEs. Additionally, developing standardized evaluation metrics for synthetic data quality encompassing distributional fidelity, utility, privacy, and computational efficiency will facilitate consistent benchmarking and foster best practices in the field. Addressing computational intensity remains a priority to ensure the model’s accessibility in resource-constrained settings, while investigating its applicability to time-series or unstructured data types could extend its utility to longitudinal studies and other healthcare applications. These efforts will solidify synthetic data’s role as a cornerstone of ethical and effective data utilization, not only in healthcare but also in other sensitive domains like finance and genomics, where similar data challenges persist. By proactively tackling limitations such as data-dependent performance and calibration trade-offs, this research lays a robust foundation for future innovations, driving transformative advancements in privacy-constrained environments.

\FloatBarrier
\section*{Code Availability}
The code used to generate results and analyses in this manuscript is available from the corresponding author upon reasonable request.  

\section*{Declaration of Generative AI and AI‐assisted Technologies in the Writing Process}

During the preparation of this work, the author(s) used \textit{ChatGPT}, an AI‐powered language model developed by OpenAI, to assist with language polishing and improving clarity. After using this tool, the author(s) reviewed and edited all generated text as needed and take(s) full responsibility for the content of the publication.

\section*{\textbf{Funding}}
Not applicable

\end{document}